\documentclass[10pt, a4paper]{article}

\usepackage{graphicx}
\usepackage{subcaption}
\usepackage{booktabs}
\usepackage{xcolor}
\usepackage{ulem}
\usepackage{todonotes}

\usepackage{lrec-coling2024} 

\newcommand{\crolaser}{\mbox{c-RoLASER}}

\title{Making Sentence Embeddings Robust to User-Generated Content}

\name{Lydia Nishimwe, Benoît Sagot, Rachel Bawden} 

\address{Inria \\
   2 rue Simone Iff, 75012 Paris, France \\
   \texttt{\{firstname.lastname\}@inria.fr}\\}

\abstract{
NLP models have been known to perform poorly on user-generated content (UGC), mainly because it presents a lot of lexical variations and deviates from the standard texts on which most of these models were trained. In this work, we focus on the robustness of LASER, a sentence embedding model, to UGC data. We evaluate this robustness by LASER's ability to represent non-standard sentences and their standard counterparts close to each other in the embedding space. Inspired by previous works extending LASER to other languages and modalities, we propose RoLASER, a robust English encoder trained using a teacher-student approach to reduce the distances between the representations of standard and UGC sentences. We show that with training only on standard and synthetic UGC-like data, RoLASER significantly improves LASER’s robustness to both natural and artificial UGC data by achieving up to $2\times$ and $11\times$ better scores. We also perform a fine-grained analysis on artificial UGC data and find that our model greatly outperforms LASER on its most challenging UGC phenomena such as keyboard typos and social media abbreviations. Evaluation on downstream tasks shows that RoLASER performs comparably to or better than LASER on standard data, while consistently outperforming it on UGC data.
\\ \newline  }

\begin{document}

\maketitleabstract

\section{Introduction}

\begin{table*}[!th]
\centering\small
\begin{tabular}{@{}p{0.12\linewidth}p{0.4\linewidth}p{0.4\linewidth}@{}}
\toprule
Corpus  & UGC sentence                                                                                                            & Standard(ised) sentence                                                                                                   \\ \midrule
MultiLexNorm\textsuperscript{$\diamond$} & {\small if i \textbf{cnt} afford the real deal , i ain't buying \textbf{nuffin} fake .. i just won't have it
}                                                                          & {\small if i \textbf{can't} afford the real deal , i ain't buying \textbf{nothing} fake .. i just won't have it
}                                                           \\ \midrule
RoCS-MT\textsuperscript{$\ddagger$} & {\small \textbf{Umm idk}, maybe \textbf{its bc} we're \textbf{DIFFERENT PEOPLE} with \textbf{DIFFERENT BODIES???}}                                                   & {\small \textbf{Um, I don't know}, maybe \textbf{it's because} we're \textbf{different people} with \textbf{different bodies?}}                                        \\  \midrule
{FLORES\textsuperscript{$\dagger$}\quad\footnotesize \texttt{abr2} + \texttt{fing} + \texttt{abr1}} & {\small \textbf{" }Luckily \textbf{nthing} happened \textbf{2} me\textbf{ ,} but I saw a macabre scene\textbf{ ,} as \textbf{ppl triwd 2} break windows in order \textbf{2 gt} out\textbf{ .}} & {\small \textbf{"}Luckily \textbf{nothing} happened \textbf{to} me\textbf{, }but I saw a macabre scene\textbf{, }as \textbf{people tried to} break windows in order \textbf{to get} out\textbf{.}} \\
\bottomrule
\end{tabular}
\caption{Example non-standard sentences from 3 different UGC corpora and their standardised versions. $\diamond$: Twitter, $\ddagger$: Reddit, $\dagger$: artificially augmented with UGC phenomena.}
\label{tab:example-sentences}
\end{table*} 

Most Natural Language Processing (NLP) models are trained on ``standard'' texts, which are edited and well written. When applied to user-generated content (UGC), these models struggle due to the high lexical variance induced by the presence of ``non-standard'' phenomena such as irregular spelling choices, evolving slang and marks of expressiveness \citep{seddah-etal-2012-french, eisenstein-2013-what, vandergoot-etal-2018-taxonomy, sanguinetti-etal-2020-treebanking}. Table~\ref{tab:example-sentences} illustrates some examples of non-standard sentences with their standardised versions. UGC has been shown to have a negative impact on NLP model performance in various tasks such as machine translation \cite{belinkov-bisk-2018-synthetic,rosalesnunez-etal-2021-understanding}, dependency parsing \citep{vandergoot-2019-indepth}, sentiment analysis \citep{kumar-etal-2020-noisy} and named entity recognition \cite{plank-etal-2020-dan}. 

\begin{figure}[!th]
\centering
\includegraphics[width=0.48\textwidth]{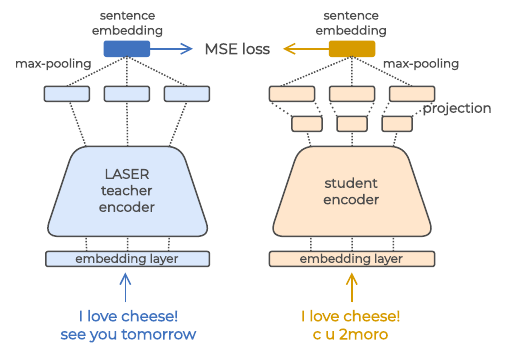}
\caption{Teacher-Student approach.}
\label{fig:distil}
\end{figure}

This performance drop of NLP models is due to their semantic vector representations (or \textit{embeddings}) not being robust to UGC, \textit{i.e.}~non-standard words and their standard counterparts do not have similar embeddings, even if they have the same meaning in the same context. Furthermore, common UGC phenomena such as acronyms (\textit{e.g.}~\textit{btw} $\rightarrow$ \textit{by the way}) and misspellings can greatly modify the tokenisation of a sentence, making it hard to represent the tokens of a UGC sentence and its normalised version in the same space. Therefore, we propose to tackle the problem at the sentence level: we consider each sentence as a whole and aim for a robust embedding that is not as affected by local, surface-level lexical variations. We frame the question of robustness to UGC as a bitext alignment problem in the sentence embedding space: \textit{how well can a sentence encoder align a standard text with its non-standard counterpart and how close are the two sentences in the embedding space?} 

Inspired by previous works extending the LASER sentence encoder \cite{artetxe-schwenk-2019-massively} to low-resource languages and the speech modality \cite{heffernan-etal-2022-bitext,duquenne-etal-2022-tmodules}, our approach is to train a student of LASER which learns to map non-standard English sentences and their standard versions close to each other in the embedding space (see Figure~\ref{fig:distil}). We compare two model architectures (one token-level and one character-aware), trained using artificially generated parallel UGC data, and we use popular bitext mining metrics for intrinsic evaluation. We also conduct an analysis of the robustness of LASER and the student models to natural and artificial UGC data in general and to each UGC phenomenon type. Finally, we analyse the performance of the models on standard data and downstream tasks such as sentence (pair) classification and semantic textual similarity.

With our robust English LASER encoder, we open the door to cross-lingual and cross-modal NLP applications on UGC data, thanks to LASER being multilingual, flexible and modular \citep{duquenne-etal-2022-tmodules}. 

Our main contributions are:
\begin{enumerate}
  \item a simple method to increase sentence-level encoder robustness to UGC by reducing the standard-UGC distance in the embedding space;
  \item RoLASER, a LASER student encoder for English more robust to natural and artificial UGC, as well as \crolaser, its character-aware equivalent;
  \item a fine-grained analysis of model robustness to artificial UGC data by UGC phenomenon type;
  \item a simple combination of data augmentation techniques for generating artificial real-life-like UGC for training and evaluation in scenarios where natural parallel UGC data is scarce. 
\end{enumerate}
We release our models and code at \url{https://github.com/lydianish/RoLASER}.

\section{Background and Related Work}

\paragraph{Language-Agnostic SEntence Representations (LASER)} One of the pioneers of large-scale multilingual sentence embedding models, LASER has known many improvements over time.
The first LASER model \citep{artetxe-schwenk-2019-massively} was a multilingual bi-LSTM \citep{schuster-paliwal-1997-bidirectional} encoder-decoder model that was trained using a machine translation objective on 93 languages, pooling the encoder's outputs to obtain a fixed-size sentence embedding. \citet{li-mak-2020-transformer} proposed T-LASER, a version of LASER built on the Transformer architecture \cite{vaswani-etal-2017-attention} and added a distance constraint to the translation loss to bring parallel sentences closer in the embedding space. After releasing LASER2, which presents some improvements with respect to the original LASER model, \citet{heffernan-etal-2022-bitext} observed that one of the major problems with it was the poor representation of low-resource languages in the multilingual sentence space. In order to tackle it, they used a teacher-student approach inspired by knowledge distillation \cite{hinton-etal-2015-distilling} to train Transformer-based encoders (student models) on monolingual and parallel xx$\rightarrow$English data to mimic the behaviour of LASER2 (the teacher model). Each of these students, called LASER3, targeted a specific low-resource language. \citet{duquenne-etal-2022-tmodules} built on this approach to build Translation Modules (T-Modules) for multilingual cross-modal translation. They trained speech and text encoders to learn from LASER2, and also trained decoders from the LASER embedding space. 
\citet{tan-etal-2023-multilingual} proposed LASER3-CO, a variant of LASER3 that integrates contrastive learning. In our work, we adapt the teacher-student approach for UGC English using a similar training setup to T-Modules, particularly the training loss.

\paragraph{Improving Model Robustness to UGC Data}
One solution to recover the performance drop of NLP models is to train or fine-tune them on UGC data. However, the scarcity of parallel annotated UGC data poses a problem. For instance, most available datasets for training or evaluating the machine translation of UGC contain only a few thousand bitexts, \textit{e.g.}~MTNT \citelanguageresource{MTNT}, PFSMB \citelanguageresource{PFSMB} and RoCS-MT \citelanguageresource{ROCSMT}. To mitigate this, data augmentation techniques have been explored to generate synthetic UGC training data. In particular, rule-based techniques consisting of character- and word-level edit operations and perturbations, as well as dictionary-based techniques, have been used to improve the robustness of NLP models to synthetic and natural UGC \cite{belinkov-bisk-2018-synthetic, karpukhin-etal-2019-training, matosveliz-etal-2019-benefits, dekker-van-der-goot-2020-synthetic, samuel-straka-2021-ufal}. In our work, we combine various types of such transformations to generate synthetic UGC data from standard data. We also analyse performance by UGC phenomenon type, similarly to \citet{rosalesnunez-etal-2021-understanding}. Data augmentation has also been used to train monolingual sentence models with a focus on improving the separation between similar and dissimilar sentences in the embedding space \cite{yan-etal-2021-consert,chuang-etal-2022-diffcse,tang-etal-2022-augcse}. Our work, however, aims to bring closer UGC sentences and their standard counterparts on the basis that they are in fact similar. Other works have also shown that character-level models can be more robust to non-standard data in such low-resource scenarios \citep{rosales-nunez-etal-2021-noisy, riabi-etal-2021-can, libovicky-etal-2022-why}, which motivates us to also explore using a student model with a character-level input embedding layer.

\section{Proposed Approach: Reducing the Standard-UGC Distance in the Embedding Space}\label{sec:approach}

We propose to train a sentence embedding model that is robust to non-standard UGC text, such that the representation assigned to non-standard text is as close as possible to its normalised equivalent without degrading model performance on standard text. We choose to work with the LASER model and aim therefore to encode non-standard text into the LASER embedding space. Although we evaluate on English in this article, this also leaves open the possibility in the future of working with other languages (for which LASER representations are also available). 

Inspired by the teacher-student approach in LASER3 \citep{heffernan-etal-2022-bitext} and T-Modules \cite{duquenne-etal-2022-tmodules}, we train a student model on standard English and UGC English data with LASER2 as the teacher (see Figure~\ref{fig:distil}). The training loss is a mean-squared error (MSE) loss, and the student model learns to minimise the distance between the two output sentence embedding vectors. As a result, it makes both standard and non-standard sentences as close as possible to the teacher's standard embeddings. This should, in theory, make it more robust to UGC phenomena. A similar approach has also been successfully applied to making monolingual sentence embeddings multilingual \cite{reimers-gurevych-2020-making}.

With LASER2 as the teacher model, we separately train two student models. The first is (BPE-based) token-level with the same architecture as RoBERTa \citep{liu-etal-2019-roberta}, which we refer to as \textbf{RoLASER} (\textbf{Ro}bust \textbf{LASER}). We also train a character-aware student  for comparison. It has a similar architecture to the first one, except for the input embedding layer, which is character-level. We refer to this model as \textbf{\crolaser}. From this point forward, LASER will be used to refer to LASER2.

Given the scarcity of natural UGC data to train such a model, we artificially generate non-standard data from standard English sentences. We achieve this by applying selected transformations from NL-Augmenter\footnote{\url{https://github.com/GEM-benchmark/NL-Augmenter}} \citep{NLAugmenter}, namely:\footnote{See Appendix~\ref{appendix:transformations} for the detailed list of transformations and random generation techniques.} 
\begin{itemize}
  \item insertion of common social media abbreviations, acronyms and slang words (\texttt{abr1}, \texttt{abr2}, \texttt{abr3}, \texttt{slng}); 
  \item contraction and expansion of auxiliary verbs (\texttt{cont}), \textit{e.g.~I am $\leftrightarrow$ I'm}, and of names of months and weekdays (\texttt{week}), \textit{e.g.~Mon. $\leftrightarrow$ Monday};
  \item insertion of misspellings such as keyboard typos or ``butter fingers'' (\texttt{fing}); homophone (\texttt{homo}) and dyslexia (\texttt{dysl}) errors, \textit{e.g.~there $\leftrightarrow$ their, lose $\leftrightarrow$ loose}; and other common spelling mistakes (\texttt{spel});
  \item visual and segmentation transformations such as Leet Speak\footnote{\url{https://en.wikipedia.org/wiki/Leet}} (\texttt{leet}), \textit{e.g.~love $\rightarrow$ l0V3}; and whitespace insertion and deletion (\texttt{spac}).
\end{itemize}

We also define a \texttt{mix\_all} transformation that randomly selects and applies a subset of the previous perturbations. For example, the last UGC sentence in Table~\ref{tab:example-sentences} was obtained via a \texttt{mix\_all} transformation which applied \texttt{abr2}, \texttt{fing} and \texttt{abr1} to a standard sentence.

\section{Evaluating Robustness}

Intuitively, the embedding space is robust if variants of the same sentence are embedded into vectors that are close to one another, \textit{i.e.}~they ideally have similar representations. However, although designed to be a semantic space, it is natural for non-semantic aspects of sentences to be represented in the space too (\textit{e.g.}~syntactic variations, language, formality, etc.), and for semantic equivalents therefore not to have identical embeddings. For the applications we envisage, our aim is for non-standard texts to be assigned embeddings that are as close as possible such that the surface form of the sentences does not impact the embeddings. To evaluate this, we use several metrics for evaluating embeddings (Section~\ref{sec:eval-metrics}) and several English normalisation-centric datasets, including both natural and artificial non-standardness (Section~\ref{sec:eval-data}).

\begin{table*}[!th]
    \centering
    \small
    \setlength\tabcolsep{2pt}
\begin{tabular*}{.9\textwidth}{@{\extracolsep{\fill}}lcc|cccccc|cc@{}}
\toprule
& \multicolumn{2}{c|}{FLORES} & \multicolumn{6}{c|}{MultiLexNorm} & \multicolumn{2}{c}{RoCS-MT} \\
& dev & devtest & \multicolumn{2}{c}{train} & \multicolumn{2}{c}{dev} & \multicolumn{2}{c|}{test} & \multicolumn{2}{c}{test} \\
Metric & std & std & std & UGC & std & UGC & std & UGC & std & UGC \\  \midrule
\# sentences & 997 & 1012 & 2360 & 2360 & 590 & 590 & 1967 & 1967 & 1922 & 1922 \\
\# tokens & 36.7k & 38.9k & 76.1k & 75.8k & 19.8k & 19.7k & 63.3k & 63.1k & 43.0k & 40.8k \\
TTR & 9.10 & 8.82 & 5.98 & 6.06 & 14.49 & 14.71 & 6.86 & 6.95 & 6.34 & 7.16 \\ 
{\scriptsize(TTR ratio)} &  &  &  & {\scriptsize(1.01)} & & {\scriptsize(1.02)} & & {\scriptsize(1.01)} &  & {\scriptsize(1.13)} \\ \bottomrule
\end{tabular*}
    \caption{Description of standard (std) and UGC data. TTR=Type-Token Ratio, TTR ratio=TTR$_{\mathrm{UGC}}$/TTR$_{\mathrm{std}}$.}
    \label{tab:eval-data-stats}
\end{table*}

\subsection{Evaluation Metrics}\label{sec:eval-metrics}

The metrics we use are pairwise cosine distance as well as xSIM and xSIM++, two metrics previously used for evaluating sentence embeddings through the proxy task of bitext mining. 

\paragraph{Average Pairwise Cosine Distance} We compute the cosine distances between the embeddings of each non-standard sentence and its normalised version and then average over all sentences in the text. For the sake of brevity, we will subsequently refer to it simply as cosine distance.

\paragraph{xSIM and xSIM++} 
Cross-lingual similarity search, or xSIM \citep{artetxe-schwenk-2019-marginbased}, is a proxy metric used for bitext mining. 
Given a set of parallel sentences in languages A (the source) and B (the target), it aligns sentences via margin-based similarity scores. It then computes the error rate of aligning each language A sentence with its language B translation from the pool of candidates (all language B sentences). xSIM++ is an extended version of the metric that discriminates better between systems and correlates more with performance on downstream tasks. It was proposed by \citet{chen-etal-2023-xsim}, who noted that xSIM was not challenging enough for many language pairs, given that the sentences in the candidate pool were often too semantically distinct (see Appendix~\ref{appendix:quantiles}). xSIM++ relies on augmenting the target set with hard negative examples, created by applying transformations that perturb the meaning of the sentences with minimal alteration to their surface form (causality alternation, number replacement and entity replacement). Note that xSIM was initially designed to be used in conjunction with the FLORES-200 dataset (see Section~\ref{sec:eval-data}) and xSIM++ only augmented the English sets (making them approximately 44 times larger). xSIM++ can therefore currently only be evaluated on xx$\rightarrow$English language pairs from FLORES-200.

\subsection{Evaluation Data}\label{sec:eval-data}

We evaluate on three English test sets representing different types of parallel non-standard data and their normalised versions.\footnote{In practice, the definition of \textit{normalised} depends on the annotation guidelines chosen.} We use two existing datasets of natural UGC (MultiLexNorm and RoCS-MT). However, in order to do a finer-grained analysis, we also create artificial UGC from FLORES-200 by applying multiple transformations. Examples from the three evaluation sets we use are provided in Table~\ref{tab:example-sentences}, and basic statistics are given in Table~\ref{tab:eval-data-stats}. Note that UGC texts tend to have fewer tokens than their standard counterparts, mainly due to the frequent use of acronyms and abbreviations.
The lexical diversity of the datasets is indicated using the type-token ratio (TTR).\footnote{The TTR is the number of unique tokens divided by the total token count; the more lexically diverse a text is, the higher the TTR. Previous work has shown that UGC texts tend to have a higher TTR due to multiple variants of the same word \citep{rosalesnunez-etal-2021-understanding}. We compute TTR based on LASER's SentencePiece tokenisation \citep{kudo-richardson-2018-sentencepiece}.} 

\paragraph{MultiLexNorm} \citeplanguageresource{MultiLexNorm2021} is a multilingual dataset created for the lexical normalisation task. We use the English subset, consisting of sentences from Twitter and their manual normalisations. The data is pretokenised and lowercased.

\paragraph{RoCS-MT} \citeplanguageresource{ROCSMT} is a multilingual dataset for the task of machine translation of UGC English into other languages: Czech (cs), German (de), French (fr), Russian (ru) and Ukranian (uk). The source sentences are from Reddit, and manual normalisations are also provided. Unlike MultiLexNorm, the data is not pretokenised nor lowercased. Casing is kept intact in the original sentences, and normalised in the standard ones.

\paragraph{FLORES-200} \citeplanguageresource{NLLB}
is a multilingual dataset consisting of parallel texts from WikiNews, WikiBooks and WikiVoyage in 200 languages. We artificially transform its English subset with UGC phenomena from NL-Augmenter as described in Section~\ref{sec:approach}. 
We subsequently refer to the original corpus as FLORES, and to the artificially augmented one as FLORES\textsuperscript{$\dagger$}.

\section{Experimental Setup}

\paragraph{Training Data} We use 2 million standard English sentences of the unshuffled deduplicated OSCAR\footnote{\url{https://huggingface.co/datasets/oscar/viewer/unshuffled_deduplicated_en}} dataset \citeplanguageresource{OSCAR}, representing 648MB of text. 
The data is split into 100 chunks of 20k sentences, each of which is artificially augmented with UGC phenomena using the \texttt{mix\_all} transformation with probability $p_{all}=0.1$ (described in Appendix~\ref{appendix:transformations}) and a different random seed, producing a 2M-sentence ``bilingual'' standard-UGC dataset. The standard sentences are passed to the teacher, while their augmented ones are passed to the student. Note that by setting a probability to apply transformations, not all sentences are augmented.\footnote{In our case, 563,343 sentences ($\approx 28.2\%$) are not transformed (see Figure~\ref{fig:distribution} in Appendix~\ref{appendix:transformations}).} Furthermore, replacement-based transformations may leave the original sentence unchanged if they find no candidate words to replace. As a result, the student model also sees standard sentences and learns to encode them (Figure~\ref{fig:distil}). 

\paragraph{Text Preprocessing} 
When fetching OSCAR data, we replace HTML line-breaking characters, do sentence splitting and filter out sentences with less than 90\% of common English characters. Afterwards, we apply the same preprocessing steps as LASER on all data, namely: removal of non-printable characters, punctuation normalisation and lowercasing \citep{artetxe-schwenk-2019-massively}. The teacher input texts are then tokenised with LASER's SentencePiece \citep{kudo-richardson-2018-sentencepiece} model (vocabulary size 50,004), and the RoLASER student inputs using RoBERTa's SentencePiece tokeniser (vocabulary size 50,265). As for the \crolaser\ student, the inputs are pretokenised on whitespace and punctuation using BERT's pretokeniser \citep{devlin-etal-2019-bert}.\footnote{\url{https://huggingface.co/google-bert/bert-base-cased}} 

\paragraph{Architectures} LASER\footnote{\url{https://github.com/facebookresearch/LASER}} is a 45M-parameter encoder with 5 bi-LSTM layers and an output embedding dimension of 1,024. RoLASER is a 108M-parameter, 12-layer Transformer encoder with 12 attention heads and a 768-output dimension, similarly to RoBERTa (without the final pooling layer). \crolaser\ is a 104M-parameter encoder with the same architecture as RoLASER except for the input embedding layer, which is a Character-CNN similar to the one used in CharacterBERT \citep{elboukkouri-etal-2020-characterbert}. Note that the students' output dimension is smaller than LASER's. Therefore, similarly to \citet{mao-nakagawa-2023-lealla}, we add a linear layer to the student encoders to project their outputs to the right size. The outputs from the teacher and students are then max-pooled to obtain sentence embedding vectors. Regarding the pooling strategy, \citet{duquenne-etal-2022-tmodules} showed that max-pooling works better than CLS-pooling for LASER students, probably because LASER itself was trained with max-pooling. While many teacher-student sentence embedding models use mean-pooling \cite{reimers-gurevych-2020-making,ham-kim-2021-semantic-alignment,mao-nakagawa-2023-lealla}, our preliminary experiments showed that max-pooling consistently performs slightly better than mean-pooling during validation. All model implementation and training are done using the Fairseq toolkit \cite{ott-2019-fairseq}. 

\paragraph{Training}
The teacher model remains frozen during training. Both student models are separately trained on 8 Tesla V100-SXM2 GPUs with a maximum number of 4,000 tokens per batch per GPU (without gradient accumulation); an Adam optimiser with parameters $\beta =(0.9,0.98)$ and $\epsilon = 10^{-6}$; learning rates of $10^{-4}$ for RoLASER and $5\times10^{-5}$ for \crolaser, both with 1,000 warm-up updates; standard, attention and activation dropouts of $0.1$; and a clip norm of $5$. Similarly to T-Modules \citep{duquenne-etal-2022-tmodules}, the training criterion is encoder similarity, and the training loss is an MSE loss with sum reduction. A checkpoint is saved every 30,000 steps. Our preliminary experiments also showed that initialising the student with a pre-trained language model performed better during validation than random initialisation. We therefore initialise RoLASER with RoBERTa,\footnote{\url{https://huggingface.co/FacebookAI/roberta-base}} and \crolaser\ with CharacterBERT.\footnote{\url{https://huggingface.co/helboukkouri/character-bert}} Table~\ref{tab:train} describes further details of the training checkpoints. 

\begin{table}[!ht]
    \centering
    \small
    \setlength\tabcolsep{0pt}
    \resizebox{\linewidth}{!}{
    \begin{tabular*}{.5\textwidth}{@{\extracolsep{\fill}}lrrrrrr@{}}
    \toprule
    Model & \#Params. & \#Epochs & \#Steps & \#Hours \\ \midrule
    RoLASER & 108M & 100 & 683k& 86 \\
    &  & \textbf{\footnotesize 98} & \textbf{\footnotesize 669k} &  \\
    c-RoLASER & 104M & 34 & 750k & 170 \\ 
    &  & \textbf{\footnotesize 32} & \textbf{\footnotesize 726k} &  \\ \bottomrule
    \end{tabular*}}
    \caption{Training details of student models. Best checkpoints are in \textbf{bold}. Trained on 8 GPUs.}
\label{tab:train}
\end{table}


\paragraph{Validation}
The best checkpoint is selected by taking the student model that minimises the MSE distance between the teacher's representation of standard text and the student's representations of (i)~standard text and (ii)~UGC text, \textit{i.e.}: 
\[
\textrm{loss}=\textrm{MSE}(L[std],m[std]) + \textrm{MSE}(L[std],m[ugc]),
\]
where $L[x]$ and $m[x]$ refer respectively to the teacher and student's representation of $x$, where $x$ can either be standard ($std$) or UGC ($ugc$) text. Framing it as a sum of two losses allows us to monitor the model's learning to minimise both distances with respect to the same anchor, using the sentence triplet $(L[std],m[std],m[ugc])$. This is different from the training loss which minimises both distances separately, \textit{i.e.}~via two separate sentence pairs $(L[std_1],m[std_1])$ and $(L[std_2],m[ugc_2])$. 
For each saved checkpoint, we compute the validation loss on the dev set of FLORES (which is also augmented with the \texttt{mix\_all} transformation) and select the checkpoint with the lowest loss.\footnote{We use a different random seed from the ones selected for augmenting the training set.} 




\section{Results and Analysis}\label{sec:results-analysis}

We evaluate LASER, RoLASER and \crolaser\ on the MultiLexNorm and RoCS-MT test sets. We also generate artificial data by applying each of the UGC transformations described in Section~\ref{sec:approach} to the standard FLORES devtest 10 times with different generation seeds, and we evaluate the models on the generated FLORES\textsuperscript{$\dagger$} sets. We first conduct an intrinsic evaluation of the student models' robustness in Section~\ref{sec:intrinsic} where we analyse whether the student models are better at representing UGC data compared to LASER, and whether they are as good as LASER on standard English. We then conduct an extrinsic evaluation in Section~\ref{sec:extrinsic} where we analyse their performance on downstream tasks such as sentence (pair) classification and semantic textual similarity. 

\subsection{Intrinsic Evaluation}
\label{sec:intrinsic}

In theory, a sentence embedding model would be robust to UGC if the cosine distance between standard and non-standard sentence pairs is small enough to ensure a perfect similarity alignment score. 
In practice, we aim to reduce cosine distances and similarity alignment error rates scores as much as possible. For each model $m$, we evaluate whether the distance between $m[ugc]$ and $m[std]$ has effectively reduced, and whether that translates into lower search error rates. We perform xSIM (and xSIM++ for FLORES) on UGC$\rightarrow$standard English bitexts. We determine the statistical significance of the student model results using an independent 2-sample t-test compared to LASER's scores.  We also compute the TTR of generated FLORES\textsuperscript{$\dagger$} files to gauge their non-standardness level, as well as their t-test compared to the TTR of the original FLORES text, and we indicate the TTR ratio with respect to the standard text. We report results on natural test sets in Section~\ref{sec:results-natural}, results on the artificial test sets for each UGC phenomenon type in Section~\ref{sec:results-by-type}, and on standard data in Section~\ref{sec:results-standard}.

\begin{table}[!th]
\centering
    \small
    \setlength\tabcolsep{8pt}
    \begin{tabular*}{0.475\textwidth}{@{\extracolsep{\fill}}lcccc@{}}
        \toprule
         & \multicolumn{2}{c}{MultiLexNorm} & \multicolumn{2}{c}{RoCS-MT} \\
        Model & cos dist & xSIM & cos dist & xSIM \\ \midrule
        LASER & 0.03 & 0.10 & 0.09 & 4.06 \\
        RoLASER & 0.02 & \textbf{0.05} & 0.06 & \textbf{2.34} \\
        {\scriptsize (improv.)} &  & {\scriptsize (2.0$\times$)} &  & {\scriptsize (1.7$\times$)} \\
        c-RoLASER & \textbf{0.01} & 0.10 & \textbf{0.05} & 3.80 \\
        {\scriptsize (improv.)} &  & {\scriptsize (1.0$\times$)} &  & {\scriptsize (1.1$\times$)} \\
        \bottomrule
    \end{tabular*}
    \caption{Cosine distance and xSIM scores on UGC$\rightarrow$standard English bitexts from natural UGC test sets. The best score for each metric is in \textbf{bold}.}
\label{tab:student-ugc-scores}
\end{table}

\subsubsection{Results on Natural UGC}
\label{sec:results-natural}

Table~\ref{tab:student-ugc-scores} illustrates the cosine distance and xSIM scores of the three models on UGC$\rightarrow$standard English bitexts from the MultiLexNorm and RoCS-MT test sets. We observe that both student models reduce the cosine distance across the board. We also note that RoCS-MT is a more challenging evaluation set as it produces much greater distances and error scores, which is consistent with it having the highest TTR ratio (Table~\ref{tab:eval-data-stats}). While RoLASER outperforms LASER with $\approx2\times$ better xSIM scores, we observe a contradictory tendency for \crolaser: despite having the lowest cosine distances, it produces minimal performance gains over LASER. We will show in Section~\ref{sec:results-standard} that this is because \crolaser\ has, on average, larger distances between its standard embeddings and LASER's.

\begin{figure}[!th]
 \centering
 \includegraphics[width=0.475\textwidth]{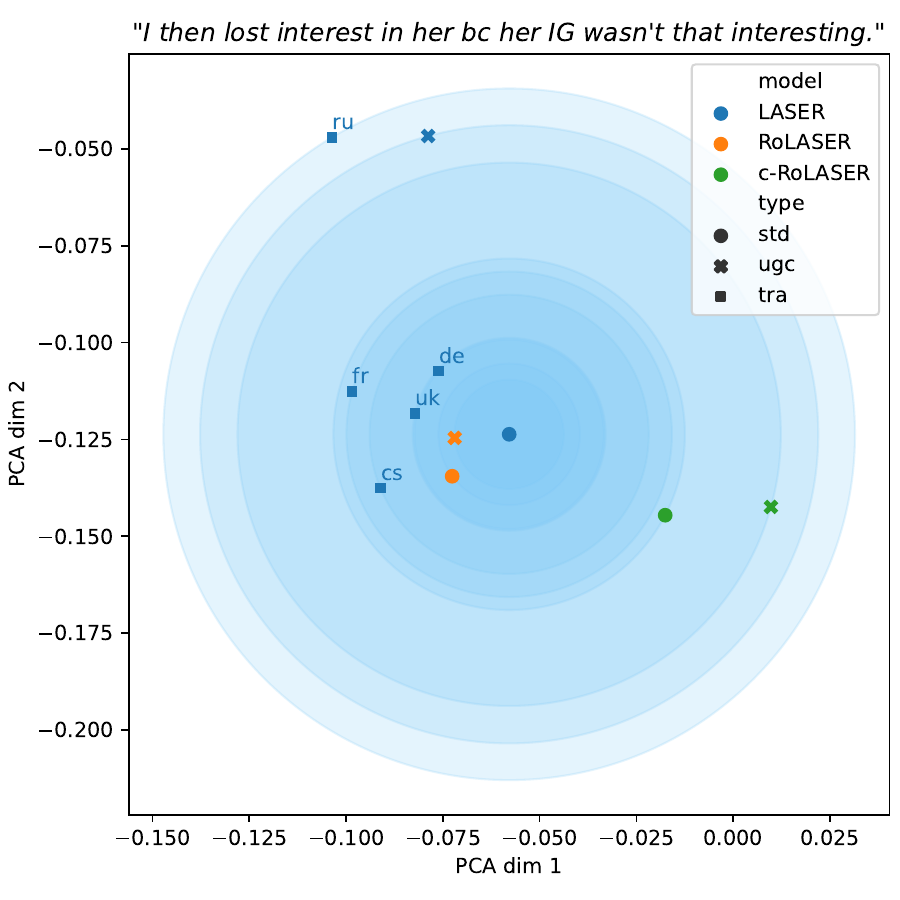}
 \caption{Visualisation of the first 2 principal components of the LASER space. The points represent the embeddings of a UGC sentence from RoCS-MT, its standardised version (std), and its translations into other languages (tra).}
 \label{fig:distances}
\end{figure}

To visually compare the students' and LASER's sentence representations, we use a 2-component PCA dimension reduction of the LASER sentence space. In Figure~\ref{fig:distances}, we plot the embeddings of the UGC sentence \textit{``I then lost interest in her bc her IG wasn't that interesting.''} from RoCS-MT,\footnote{We choose this example because it illustrates the trends observed on the RoCS-MT test set.} its normalised version \textit{``I then lost interest in her, because her Instagram wasn’t that interesting.''}, and its translations in five other languages. We evaluate the distance preservation in the reduced dimensions and obtain a Spearman's correlation of $r=0.69$ between Euclidean distances in the reduced and original space. We observe that both RoLASER and \crolaser\ have a shorter standard-UGC distance than LASER. Furthermore, RoLASER's standard and UGC embeddings are closer to LASER's than any of the other languages. However, \crolaser's standard embedding remains far from LASER's, which explains its poor xSIM scores. 

\subsubsection{Results by UGC Phenomenon Type}
\label{sec:results-by-type}

\begin{table*}[!ht]
    \centering
    \small
    \setlength\tabcolsep{0pt}
    \begin{tabular*}{\textwidth}{@{\extracolsep{\fill}}lcccccccccccccc@{}}
    \toprule
    UGC type & abr1 & abr2 & abr3 & cont & dysl & fing & homo & leet & slng & spac & spel & week & mix\_all \\ 
    {\scriptsize (TTR ratio)} & {\scriptsize (0.93**)} & {\scriptsize (0.98**)} & {\scriptsize (1.00**)} & {\scriptsize (1.00**)} & {\scriptsize (0.99**)} & {\scriptsize (1.05**)} & {\scriptsize (0.98**)} & {\scriptsize (0.76**)} & {\scriptsize (0.98**)} & {\scriptsize (1.01*)} & {\scriptsize (1.02**)} & {\scriptsize (1.00**)} & {\scriptsize (1.01**)} \\ \midrule
    \multicolumn{14}{c}{\textit{Average pairwise cosine distance}} \\
    LASER & 0.03 & 0.08 & \textbf{0.00} & \textbf{0.00} & 0.04 & 0.07 & 0.05 & 0.22 & 0.02 &  0.09 & 0.04 & \textbf{0.00} & 0.05 \\
    RoLASER & \textbf{0.00}** & \textbf{0.00}** & \textbf{0.00} & \textbf{0.00} & \textbf{0.00}** & 0.02** & \textbf{0.00}** & 0.01** & \textbf{0.00}** & \textbf{0.01}** & \textbf{0.01}** & \textbf{0.00} & \textbf{0.00}** \\
   c-RoLASER & \textbf{0.00}** & 0.01** & \textbf{0.00} & \textbf{0.00} & 0.01** & \textbf{0.01}** & 0.01** &  \textbf{0.00}** & 0.01** & 0.03** & \textbf{0.01}** & \textbf{0.00} & 0.01** \\
   \midrule
   \multicolumn{14}{c}{\textit{xSIM++}} \\
    LASER & 4.01 & 15.81 & \textbf{0.10} & \textbf{0.20} & 5.83 & 19.60 & 10.13 & 68.60 & 2.39 & 22.76 & 7.85 & 2.08 & 13.17 \\
    RoLASER & \textbf{0.75}* & \textbf{1.58}** & 0.40 & 0.40 & \textbf{0.79}** & \textbf{7.09}* & \textbf{0.96}** & \textbf{3.03}** & \textbf{0.61}** & \textbf{2.45}** & \textbf{2.14}** & \textbf{0.40}** & \textbf{1.22}** \\
    {\scriptsize (improv.)} & {\scriptsize (5.4$\times$)} & {\scriptsize (10.0$\times$)} &  &  & {\scriptsize (7.4$\times$)} & {\scriptsize (2.8$\times$)} & {\scriptsize (10.6$\times$)} & {\scriptsize (22.7$\times$)} & {\scriptsize (3.9$\times$)} & {\scriptsize (9.3$\times$)} & {\scriptsize (3.7$\times$)} & {\scriptsize (5.2$\times$)} & {\scriptsize (10.8$\times$)} \\ 
    c-RoLASER & 13.16 & 19.37 & 11.76 & 12.94 & 15.71 & 13.49 & 16.93 & 12.56** & 14.54 & 19.72 & 15.00 & 11.46 & 15.74 \\
    {\scriptsize (improv.)} & & &  &  & & {\scriptsize (1.5$\times$)} &  & {\scriptsize (5.5$\times$)} & & {\scriptsize (1.2$\times$)} & & & & \\ 
    \bottomrule
    \end{tabular*}
    \caption{Cosine distance and xSIM++ scores for all models on UGC$\rightarrow$standard English bitext from each UGC type of FLORES\textsuperscript{$\dagger$} devtest, averaged across 10 data generation seeds. The best score for each type is in \textbf{bold}. *: $p<0.05$, **: $p<0.001$, statistical significance with respect to LASER's scores.}
    \label{tab:artificial-ugc-student-xsimpp}
\end{table*}

\begin{table*}[!th]
    \centering
    \small
    \setlength\tabcolsep{2pt}
\begin{tabular*}{\textwidth}{@{\extracolsep{\fill}}lcccccccccc@{}}
\toprule

& \multicolumn{2}{c}{cs$\rightarrow$en} & \multicolumn{2}{c}{de$\rightarrow$en} & \multicolumn{2}{c}{fr$\rightarrow$en} & \multicolumn{2}{c}{ru$\rightarrow$en} & \multicolumn{2}{c}{uk$\rightarrow$en} \\ 
Model [en]  & UGC & std & UGC & std & UGC & std & UGC & std & UGC & std \\ \midrule
LASER & 9.11 & \textbf{3.28} & 6.56 & 0.83 & 10.20 & \textbf{4.68} & 11.76 & 5.93 & 8.79 & 2.39 \\
RoLASER & \textbf{7.23} & 3.33 & \textbf{4.94} & \textbf{0.73} & \textbf{9.21} & 4.79 & \textbf{10.15} & \textbf{4.89} & \textbf{6.61} & \textbf{2.34} \\
c-RoLASER & 13.94  & 7.49 & 9.78 & 4.37 & 15.04 & 9.31 & 16.44 & 10.87 & 13.53 & 7.13 \\ \midrule
& \multicolumn{2}{c}{en$\rightarrow$cs} & \multicolumn{2}{c}{en$\rightarrow$de} & \multicolumn{2}{c}{en$\rightarrow$fr} & \multicolumn{2}{c}{en$\rightarrow$ru} & \multicolumn{2}{c}{en$\rightarrow$uk} \\ 
Model [en] & UGC & std & UGC & std & UGC & std & UGC & std & UGC & std \\ \midrule
LASER & 9.11 & \textbf{2.71} & 5.83 & \textbf{0.57} & 10.87  & \textbf{5.10} & 11.71 & \textbf{5.88} & 8.79 & 2.55  \\
RoLASER & \textbf{7.02} & 3.17 & \textbf{4.58} & 0.83 & \textbf{8.64} & 5.20 & \textbf{10.20} & 6.35 & \textbf{6.04} & \textbf{2.34} \\
c-RoLASER & 18.26 & 9.16 & 13.16 & 5.52 & 19.25 & 11.76 & 23.31 & 14.36 & 17.74 & 8.58 \\
\bottomrule
\end{tabular*}
\caption{xSIM scores on xx$\rightarrow$English and English$\rightarrow$xx bitexts from RoCS-MT. The results compare all models for embedding UGC and standard (std) English. Only LASER is used to embed the non-English languages. The best score for each language pair is in \textbf{bold}.}
\label{tab:multilingual-rocsmt}
\end{table*}

Table~\ref{tab:artificial-ugc-student-xsimpp} illustrates the cosine distance and the xSIM++ scores of the three models on the FLORES\textsuperscript{$\dagger$} devtest for all UGC types, as well as the ratio of TTRs of the UGC texts with respect to the standard text.\footnote{See Figure~\ref{fig:ttr} in Appendix~\ref{appendix:transformations} for more interpretation of the TTR ratios.} We report the xSIM results in Table~\ref{tab:artificial-ugc-student-xsim} (Appendix~\ref{appendix:xsim}).
All of LASER's xSIM++ scores are (highly) significantly\footnote{significant: $p<0.05$, highly significant: $p<0.001$.} different from zero (the expected mean), suggesting a lack of robustness of LASER to artificial UGC types.

We observe that both student models have highly significantly reduced the cosine distances to close to zero. We also note that cosine distance scores for LASER on most UGC types are less than $0.07$, which is the minimum, or 0\textsuperscript{th} percentile, for LASER on all xx$\rightarrow$English FLORES language pairs (see Figure~\ref{fig:cos-quantiles} in Appendix~\ref{appendix:quantiles}). In other words, LASER mostly represents UGC English closer to standard English than it does all the other languages, which is reasonable considering UGC English is still English. The UGC type that it embeds the furthest from standard English is \texttt{leet} with a cosine distance of $0.22$,  which is in the 35\textsuperscript{th} percentile. This means LASER considers that 35\% of FLORES languages are closer to standard English than Leet Speak English.

With xSIM++, the three most challenging transformations for LASER are \texttt{leet}, \texttt{space} and \texttt{fing}. Intuitively, they are the ones that ``shatter'' subword tokenisation the most because they perform character-level perturbations. In fact, \texttt{leet} and \texttt{fing} have the lowest and highest TTR ratios respectively. The next batch of challenging transformations apply more word-level perturbations (\texttt{abr2} and \texttt{homo}). Other noteworthy transformations are the ones with very low xSIM++ scores and cosine distances of zero: \texttt{abr3}, \texttt{cont}, \texttt{week}. Finally, the results suggest that \texttt{mix\_all} is challenging enough to be a good attempt at generating comprehensive, real-life-like artificial UGC. 

RoLASER outperforms LASER artificial UGC (as shown by the $10.8\times$ better xSIM++ score on \texttt{mix\_all}). We also observe major performance gains for several UGC transformations: $22.7\times$ better for \texttt{leet}, and between $3.9\times$ and $10.6\times$ for most of the other types. \texttt{fing} remains the most challenging one for RoLASER as it obtains the highest cosine distance and xSIM++ score (which is still $2.8\times$ better than LASER's). 
Lastly, RoLASER slightly degrades LASER's performance on \texttt{cont} and \texttt{abr3}. This is likely because these phenomena are already frequent in standard data, which means that the original LASER has already been trained to deal with them efficiently.  It could also be that they perform minimal perturbations on the original text (as shown by their TTR ratio of $1.00$). 

We also note that all the RoLASER xSIM++ scores are less than $7.21\%$, which is the minimum score for LASER on all xx$\rightarrow$English FLORES language pairs (see Figure~\ref{fig:quantiles} in Appendix~\ref{appendix:quantiles}). It is akin to saying that RoLASER aligns UGC English to standard English better than LASER does all the other languages.

However, the \crolaser\ results are disappointing: it degrades the performance on all types, except for \texttt{leet}, \texttt{fing} and \texttt{spac}, and it never outperforms RoLASER. This is consistent with the results on natural UGC that \crolaser\ struggles to map its standard embeddings to LASER's.

\subsubsection{Results on Standard Data}
\label{sec:results-standard}

It is also important to evaluate whether the student models' reduced UGC-standard distances introduce a performance drop on standard data. In theory, this should not be the case since they are also trained to minimise the distance between their standard embeddings $m[std]$ on the one hand, and LASER's standard embeddings $L[std]$ on the other. 

We evaluate all models on the task of bitext alignment on the five xx-English language pairs of RoCS-MT. Table~\ref{tab:multilingual-rocsmt} shows the xSIM scores in both xx$\rightarrow$English and English$\rightarrow$xx directions,\footnote{xSIM is not symmetrical: scores are not comparable across both language pair directions \cite{chen-etal-2023-xsim}.} where English is either UGC or standard (std). LASER is used to embed all non-English sentences, while both LASER and the student models are used for the English sentences.

As is expected, standard English consistently produces better results than UGC for all the models. We also observe that RoLASER improves on LASER's performance for standard English in the xx$\rightarrow$English direction. This is likely because the student specialised in standard English as the target language during training. In the English$\rightarrow$xx  direction however, RoLASER only surpasses LASER about half the time. As for UGC English, we observe that RoLASER produces the best results in both directions, while \crolaser\ degrades LASER's performance. 

\begin{table}[!ht]
    \centering
    \small
    \setlength\tabcolsep{2pt}
    \begin{tabular*}{0.43\textwidth}{@{\extracolsep{\fill}}lrrr@{}}
        \toprule
        Model & FLORES & MultiLexNorm & RoCS-MT \\ \midrule
        RoLASER & 0.02 & 0.04 & 0.05 \\
        c-RoLASER & 0.05 & 0.09 & 0.13 \\
        \bottomrule
    \end{tabular*}
    \caption{Cosine distance between the students' and LASER's standard embeddings.}
\label{tab:student-std-distances}
\end{table}

To better understand these results, we compare the standard English embeddings from the student models with LASER's on all test sets.  We illustrate in Table~\ref{tab:student-std-distances} the average pairwise cosine distance between them. They show that RoLASER has managed to effectively minimise the distance between its standard embeddings and LASER's, which manifests as performance gains observed in the bilingual alignments (Table~\ref{tab:multilingual-rocsmt}). 
However, \crolaser\ struggles to map its standard embeddings to LASER's, especially on RoCS-MT. This explains its poor performance in general. In other words, \crolaser\ has successfully reduced the distance between its UGC and standard embeddings to almost zero (see Table~\ref{tab:artificial-ugc-student-xsimpp}), but it lags behind when bridging the gap between its standard embeddings and LASER's. One reason for this could be that character-level tokenisation results in very long sequences, making it a difficult task for the model pool their representations into one fixed-sized vector capturing all semantic information. Nonetheless, we suspect that \crolaser\ could benefit from longer and better optimised training.

\subsection{Extrinsic Evaluation}
\label{sec:extrinsic}

To support the results of the intrinsic evaluation (Section~\ref{sec:intrinsic}), we evaluate our models' performance on downstream tasks from MTEB, the Massive Text Embedding Benchmark \citeplanguageresource{MTEB}. We select four tasks spanning three types:
\begin{enumerate}
    \item \textbf{Sentence classification}, which predicts labels from sentence embeddings, \textit{e.g.}~sentiment labels: \texttt{TweetSentimentExtractionClassification} (T-SentExt).
    \item \textbf{Sentence pair classification}, which predicts a binary label from sentence embeddings, \textit{e.g.}~whether two sentences are paraphrases: \texttt{TwitterSemEval2015} (T-SemEval) and \texttt{TwitterURLCorpus} (T-URL).
    \item \textbf{Semantic textual similarity}, which examines the degree of semantic equivalence between two sentences: \texttt{STSBenchmark} (STS).
\end{enumerate}
Note that the first three tasks are evaluated on UGC, specifically Twitter data. The last one is evaluated on more standard texts from image captions, news headlines and user forums.

\begin{table}[!ht]
    \centering
    \small
    \setlength\tabcolsep{0pt}
    \begin{tabular*}{.475\textwidth}{@{\extracolsep{\fill}}lrrrrr@{}}
    \toprule
    Model & T-SentExt\textsuperscript{$\diamond$} & T-SemEval\textsuperscript{$\dagger$} & T-URL\textsuperscript{$\dagger$} & STS\textsuperscript{$\ddagger$}  \\ \midrule
    LASER & 50.64 & 59.57 & 81.48 & \textbf{69.77} \\
    RoLASER & \textbf{51.96} & \textbf{60.68} & \textbf{81.79} & 69.61 \\
    c-RoLASER & 49.29 & 55.32 & 76.80 & 68.13 \\ \bottomrule
    \end{tabular*}
    \caption{Scores (\%) on 4 MTEB tasks. The best score for each metric is in \textbf{bold}. $\diamond$: accuracy, $\dagger$: average precision on cosine similarity,  $\ddagger$: Spearman's correlation on cosine similarity.}
    \label{tab:mteb}
\end{table} 

Table~\ref{tab:mteb} shows the scores of our models on the four tasks (along with their corresponding evaluation metrics). RoLASER consistently outperforms LASER on the first three tasks on Twitter data, while it is almost as good as LASER on the standard STS task. This is in agreement with our findings in  Section~\ref{sec:intrinsic} that RoLASER is better than LASER at encoding non-standard data and achieves comparable performance on standard data. On the other hand, c-RoLASER remains the worst across all tasks and greatly degrades LASER's performance.

\section{Conclusion}

In this work, we frame the question of LASER's robustness to UGC as a bitext alignment problem where we aim to align standard sentences and their non-standard equivalents. We propose RoLASER, a Transformer-based encoder student of LASER, trained with the objective of minimising the distances between standard and non-standard sentence pairs in the embedding space. The model is trained solely on standard and synthetic UGC-like English data. We also consider a character-aware student, \crolaser,  and find that the token-level RoLASER performs best overall while the \crolaser\ struggles to map its standard embeddings to LASER's. 

We find that RoLASER is significantly more robust than LASER on natural UGC, achieving up to $2\times$ better xSIM scores. We also evaluate it on standard data and downstream tasks and show that it improves, or at least matches, LASER's performance. Furthermore, we perform a fine-grained analysis of the models' robustness with respect to artificially generated data by type of UGC phenomena. We show that RoLASER achieves roughly $11\times$ better xSIM++ scores than LASER on artificial UGC, and up to $23\times$ better on Leet Speak, the most difficult UGC type for LASER. We also find that the most challenging phenomena are those with character-level perturbations that shatter subword tokenisation.

For future work, we plan to extend RoLASER to more languages and their corresponding UGC phenomena. We will also consider ways to improve \crolaser, such as using a thin-deep architecture \cite{tay-etal-2021-charformer}, or a token-level model with a small enough vocabulary size to be close to the character level.

\section{Limitations}

The ambiguity introduced by non-standard words in language could be problematic. For example, \textit{smh} could mean \textit{shaking my head} or \textit{so much hate}, and our approach would try to map both to the same space. One way to resolve this ambiguity would be to use the surrounding sentences as context. Though it is an interesting line of research to pursue, it is outside the scope of this article. Thankfully, such cases are rare and the model has proved to do well in general across multiple UGC types.

There is also a possible domain mismatch between the type of data used to train the our models and the data on which we test.  RoLASER is trained and validated on standard data artificially augmented with UGC phenomena and is evaluated on (scarce) parallel UGC data from social media. However, the results show that the model is able to generalise well on natural UGC data without having been trained or fine-tuned on it.

\section{Acknowledgements}
We thank the anonymous reviewers for their constructive feedback, and Paul-Ambroise Duquenne for his insights on LASER. This work was granted access to the HPC resources of IDRIS under the allocations 2023-AD011012254R1 and 2023-AD011013674R1 made by GENCI. This work was funded by the last two authors' chairs in the PRAIRIE institute funded by the French national agency ANR as part of the ``Investissements d'avenir'' programme under the reference ANR-19-P3IA-0001.

\nocite{*}

\section{Bibliographical References}
\label{sec:reference}
\bibliographystyle{lrec-coling2024-natbib}

\begin{thebibliography}{40}
\expandafter\ifx\csname natexlab\endcsname\relax\def\natexlab#1{#1}\fi

\bibitem[{Artetxe and Schwenk(2019{\natexlab{a}})}]{artetxe-schwenk-2019-marginbased}
Mikel Artetxe and Holger Schwenk. 2019{\natexlab{a}}.
\newblock \href {https://doi.org/10.18653/v1/P19-1309} {Margin-based {{Parallel Corpus Mining}} with {{Multilingual Sentence Embeddings}}}.
\newblock In \emph{Proceedings of the 57th {{Annual Meeting}} of the {{Association}} for {{Computational Linguistics}}}, pages 3197--3203, {Florence, Italy}. {Association for Computational Linguistics}.

\bibitem[{Artetxe and Schwenk(2019{\natexlab{b}})}]{artetxe-schwenk-2019-massively}
Mikel Artetxe and Holger Schwenk. 2019{\natexlab{b}}.
\newblock \href {https://doi.org/10.1162/tacl_a_00288} {Massively {{Multilingual Sentence Embeddings}} for {{Zero-Shot Cross-Lingual Transfer}} and {{Beyond}}}.
\newblock In \emph{Transactions of the {{Association}} for {{Computational Linguistics}}}, volume~7, pages 597--610, {Cambridge, MA}. {MIT Press}.

\bibitem[{Belinkov and Bisk(2018)}]{belinkov-bisk-2018-synthetic}
Yonatan Belinkov and Yonatan Bisk. 2018.
\newblock \href {https://openreview.net/forum?id=BJ8vJebC-} {Synthetic and natural noise both break neural machine translation}.
\newblock In \emph{Proceedings of the 6th International Conference on Learning Representations}, Vancouver, BC, Canada.

\bibitem[{Chen et~al.(2023)Chen, Heffernan, {\c{C}}elebi, Mourachko, and Schwenk}]{chen-etal-2023-xsim}
Mingda Chen, Kevin Heffernan, Onur {\c{C}}elebi, Alexandre Mourachko, and Holger Schwenk. 2023.
\newblock \href {https://doi.org/10.18653/v1/2023.acl-short.10} {x{SIM}++: An improved proxy to bitext mining performance for low-resource languages}.
\newblock In \emph{Proceedings of the 61st Annual Meeting of the Association for Computational Linguistics (Volume 2: Short Papers)}, pages 101--109, Toronto, Canada. Association for Computational Linguistics.

\bibitem[{Chuang et~al.(2022)Chuang, Dangovski, Luo, Zhang, Chang, Soljacic, Li, Yih, Kim, and Glass}]{chuang-etal-2022-diffcse}
Yung-Sung Chuang, Rumen Dangovski, Hongyin Luo, Yang Zhang, Shiyu Chang, Marin Soljacic, Shang-Wen Li, Scott Yih, Yoon Kim, and James Glass. 2022.
\newblock \href {https://doi.org/10.18653/v1/2022.naacl-main.311} {{D}iff{CSE}: Difference-based contrastive learning for sentence embeddings}.
\newblock In \emph{Proceedings of the 2022 Conference of the North American Chapter of the Association for Computational Linguistics: Human Language Technologies}, pages 4207--4218, Seattle, United States. Association for Computational Linguistics.

\bibitem[{Dekker and van~der Goot(2020)}]{dekker-van-der-goot-2020-synthetic}
Kelly Dekker and Rob van~der Goot. 2020.
\newblock \href {https://aclanthology.org/2020.lrec-1.773} {Synthetic data for {E}nglish lexical normalization: How close can we get to manually annotated data?}
\newblock In \emph{Proceedings of the Twelfth Language Resources and Evaluation Conference}, pages 6300--6309, Marseille, France. European Language Resources Association.

\bibitem[{Devlin et~al.(2019)Devlin, Chang, Lee, and Toutanova}]{devlin-etal-2019-bert}
Jacob Devlin, Ming-Wei Chang, Kenton Lee, and Kristina Toutanova. 2019.
\newblock \href {https://doi.org/10.18653/v1/N19-1423} {{BERT}: Pre-training of deep bidirectional transformers for language understanding}.
\newblock In \emph{Proceedings of the 2019 Conference of the North {A}merican Chapter of the Association for Computational Linguistics: Human Language Technologies, Volume 1 (Long and Short Papers)}, pages 4171--4186, Minneapolis, Minnesota. Association for Computational Linguistics.

\bibitem[{Dhole et~al.(2021)Dhole, Gangal, Gehrmann, Gupta, Li, Mahamood, Mahendiran, Mille, Srivastava, Tan, Wu, Sohl-Dickstein, Choi, Hovy, Dusek, Ruder, Anand, Aneja, Banjade, Barthe, Behnke, Berlot-Attwell, Boyle, Brun, Cabezudo, Cahyawijaya, Chapuis, Che, Choudhary, Clauss, Colombo, Cornell, Dagan, Das, Dixit, Dopierre, Dray, Dubey, Ekeinhor, Giovanni, Gupta, Gupta, Hamla, Han, Harel-Canada, Honore, Jindal, Joniak, Kleyko, Kovatchev, Krishna, Kumar, Langer, Lee, Levinson, Liang, Liang, Liu, Lukyanenko, Marivate, de~Melo, Meoni, Meyer, Mir, Moosavi, Muennighoff, Mun, Murray, Namysl, Obedkova, Oli, Pasricha, Pfister, Plant, Prabhu, Pais, Qin, Raji, Rajpoot, Raunak, Rinberg, Roberts, Rodriguez, Roux, S., Sai, Schmidt, Scialom, Sefara, Shamsi, Shen, Shi, Shi, Shvets, Siegel, Sileo, Simon, Singh, Sitelew, Soni, Sorensen, Soto, Srivastava, Srivatsa, Sun, T, Tabassum, Tan, Teehan, Tiwari, Tolkiehn, Wang, Wang, Wang, Wang, Wei, Wilie, Winata, Wu, Wydmański, Xie, Yaseen, Yee, Zhang, and Zhang}]{NLAugmenter}
Kaustubh~D. Dhole, Varun Gangal, Sebastian Gehrmann, Aadesh Gupta, Zhenhao Li, Saad Mahamood, Abinaya Mahendiran, Simon Mille, Ashish Srivastava, Samson Tan, Tongshuang Wu, Jascha Sohl-Dickstein, Jinho~D. Choi, Eduard Hovy, Ondrej Dusek, Sebastian Ruder, Sajant Anand, Nagender Aneja, Rabin Banjade, Lisa Barthe, Hanna Behnke, Ian Berlot-Attwell, Connor Boyle, Caroline Brun, Marco Antonio~Sobrevilla Cabezudo, Samuel Cahyawijaya, Emile Chapuis, Wanxiang Che, Mukund Choudhary, Christian Clauss, Pierre Colombo, Filip Cornell, Gautier Dagan, Mayukh Das, Tanay Dixit, Thomas Dopierre, Paul-Alexis Dray, Suchitra Dubey, Tatiana Ekeinhor, Marco~Di Giovanni, Rishabh Gupta, Rishabh Gupta, Louanes Hamla, Sang Han, Fabrice Harel-Canada, Antoine Honore, Ishan Jindal, Przemyslaw~K. Joniak, Denis Kleyko, Venelin Kovatchev, Kalpesh Krishna, Ashutosh Kumar, Stefan Langer, Seungjae~Ryan Lee, Corey~James Levinson, Hualou Liang, Kaizhao Liang, Zhexiong Liu, Andrey Lukyanenko, Vukosi Marivate, Gerard de~Melo, Simon Meoni, Maxime
  Meyer, Afnan Mir, Nafise~Sadat Moosavi, Niklas Muennighoff, Timothy Sum~Hon Mun, Kenton Murray, Marcin Namysl, Maria Obedkova, Priti Oli, Nivranshu Pasricha, Jan Pfister, Richard Plant, Vinay Prabhu, Vasile Pais, Libo Qin, Shahab Raji, Pawan~Kumar Rajpoot, Vikas Raunak, Roy Rinberg, Nicolas Roberts, Juan~Diego Rodriguez, Claude Roux, Vasconcellos P.~H. S., Ananya~B. Sai, Robin~M. Schmidt, Thomas Scialom, Tshephisho Sefara, Saqib~N. Shamsi, Xudong Shen, Haoyue Shi, Yiwen Shi, Anna Shvets, Nick Siegel, Damien Sileo, Jamie Simon, Chandan Singh, Roman Sitelew, Priyank Soni, Taylor Sorensen, William Soto, Aman Srivastava, KV~Aditya Srivatsa, Tony Sun, Mukund~Varma T, A~Tabassum, Fiona~Anting Tan, Ryan Teehan, Mo~Tiwari, Marie Tolkiehn, Athena Wang, Zijian Wang, Gloria Wang, Zijie~J. Wang, Fuxuan Wei, Bryan Wilie, Genta~Indra Winata, Xinyi Wu, Witold Wydmański, Tianbao Xie, Usama Yaseen, M.~Yee, Jing Zhang, and Yue Zhang. 2021.
\newblock \href {http://arxiv.org/abs/2112.02721} {{NL-Augmenter: A Framework for Task-Sensitive Natural Language Augmentation}}.

\bibitem[{Duquenne et~al.(2022)Duquenne, Gong, Sagot, and Schwenk}]{duquenne-etal-2022-tmodules}
Paul-Ambroise Duquenne, Hongyu Gong, Beno{\^i}t Sagot, and Holger Schwenk. 2022.
\newblock \href {https://aclanthology.org/2022.emnlp-main.391} {T-{{Modules}}: {{Translation Modules}} for {{Zero-Shot Cross-Modal Machine Translation}}}.
\newblock In \emph{Proceedings of the 2022 {{Conference}} on {{Empirical Methods}} in {{Natural Language Processing}}}, pages 5794--5806, {Abu Dhabi, United Arab Emirates}. {Association for Computational Linguistics}.

\bibitem[{Eisenstein(2013)}]{eisenstein-2013-what}
Jacob Eisenstein. 2013.
\newblock \href {https://aclanthology.org/N13-1037} {What to do about bad language on the internet}.
\newblock In \emph{Proceedings of the 2013 {{Conference}} of the {{North American Chapter}} of the {{Association}} for {{Computational Linguistics}}: {{Human Language Technologies}}}, pages 359--369, {Atlanta, Georgia}. {Association for Computational Linguistics}.

\bibitem[{El~Boukkouri et~al.(2020)El~Boukkouri, Ferret, Lavergne, Noji, Zweigenbaum, and Tsujii}]{elboukkouri-etal-2020-characterbert}
Hicham El~Boukkouri, Olivier Ferret, Thomas Lavergne, Hiroshi Noji, Pierre Zweigenbaum, and Jun'ichi Tsujii. 2020.
\newblock \href {https://doi.org/10.18653/v1/2020.coling-main.609} {{{CharacterBERT}}: {{Reconciling ELMo}} and {{BERT}} for {{Word-Level Open-Vocabulary Representations From Characters}}}.
\newblock In \emph{Proceedings of the 28th {{International Conference}} on {{Computational Linguistics}}}, pages 6903--6915, {Barcelona, Spain (Online)}. {International Committee on Computational Linguistics}.

\bibitem[{Ham and Kim(2021)}]{ham-kim-2021-semantic-alignment}
Jiyeon Ham and Eun-Sol Kim. 2021.
\newblock \href {https://doi.org/10.18653/v1/2021.findings-emnlp.153} {Semantic alignment with calibrated similarity for multilingual sentence embedding}.
\newblock In \emph{Findings of the Association for Computational Linguistics: EMNLP 2021}, pages 1781--1791, Punta Cana, Dominican Republic. Association for Computational Linguistics.

\bibitem[{Heffernan et~al.(2022)Heffernan, {\c C}elebi, and Schwenk}]{heffernan-etal-2022-bitext}
Kevin Heffernan, Onur {\c C}elebi, and Holger Schwenk. 2022.
\newblock \href {https://aclanthology.org/2022.findings-emnlp.154} {Bitext {{Mining Using Distilled Sentence Representations}} for {{Low-Resource Languages}}}.
\newblock In \emph{Findings of the {{Association}} for {{Computational Linguistics}}: {{EMNLP}} 2022}, pages 2101--2112, {Abu Dhabi, United Arab Emirates}. {Association for Computational Linguistics}.

\bibitem[{Hinton et~al.(2015)Hinton, Vinyals, and Dean}]{hinton-etal-2015-distilling}
Geoffrey Hinton, Oriol Vinyals, and Jeff Dean. 2015.
\newblock \href {http://arxiv.org/abs/1503.02531} {Distilling the knowledge in a neural network}.
\newblock In \emph{Proceedings of the 2014 NIPS Deep Learning and Representation Learning Workshop}, Montreal, Quebec, Canada.

\bibitem[{Karpukhin et~al.(2019)Karpukhin, Levy, Eisenstein, and Ghazvininejad}]{karpukhin-etal-2019-training}
Vladimir Karpukhin, Omer Levy, Jacob Eisenstein, and Marjan Ghazvininejad. 2019.
\newblock \href {https://doi.org/10.18653/v1/D19-5506} {Training on {{Synthetic Noise Improves Robustness}} to {{Natural Noise}} in {{Machine Translation}}}.
\newblock In \emph{Proceedings of the 5th {{Workshop}} on {{Noisy User-generated Text}} ({{W-NUT}} 2019)}, pages 42--47, {Hong Kong, China}. {Association for Computational Linguistics}.

\bibitem[{Kudo and Richardson(2018)}]{kudo-richardson-2018-sentencepiece}
Taku Kudo and John Richardson. 2018.
\newblock \href {https://doi.org/10.18653/v1/D18-2012} {{S}entence{P}iece: A simple and language independent subword tokenizer and detokenizer for neural text processing}.
\newblock In \emph{Proceedings of the 2018 Conference on Empirical Methods in Natural Language Processing: System Demonstrations}, pages 66--71, Brussels, Belgium. Association for Computational Linguistics.

\bibitem[{Kumar et~al.(2020)Kumar, Makhija, and Gupta}]{kumar-etal-2020-noisy}
Ankit Kumar, Piyush Makhija, and Anuj Gupta. 2020.
\newblock \href {https://doi.org/10.18653/v1/2020.wnut-1.3} {Noisy {{Text Data}}: {{Achilles}}' {{Heel}} of {{BERT}}}.
\newblock In \emph{Proceedings of the {{Sixth Workshop}} on {{Noisy User-generated Text}} ({{W-NUT}} 2020)}, pages 16--21, {Online}. {Association for Computational Linguistics}.

\bibitem[{Li and Mak(2020)}]{li-mak-2020-transformer}
Wei Li and Brian Mak. 2020.
\newblock \href {http://arxiv.org/abs/2008.08567} {Transformer based multilingual document embedding model}.
\newblock \emph{CoRR}, abs/2008.08567.

\bibitem[{Libovick{\'y} et~al.(2022)Libovick{\'y}, Schmid, and Fraser}]{libovicky-etal-2022-why}
Jind{\v{r}}ich Libovick{\'y}, Helmut Schmid, and Alexander Fraser. 2022.
\newblock \href {https://doi.org/10.18653/v1/2022.findings-acl.194} {Why don{'}t people use character-level machine translation?}
\newblock In \emph{Findings of the Association for Computational Linguistics: ACL 2022}, pages 2470--2485, Dublin, Ireland. Association for Computational Linguistics.

\bibitem[{Liu et~al.(2019)Liu, Ott, Goyal, Du, Joshi, Chen, Levy, Lewis, Zettlemoyer, and Stoyanov}]{liu-etal-2019-roberta}
Yinhan Liu, Myle Ott, Naman Goyal, Jingfei Du, Mandar Joshi, Danqi Chen, Omer Levy, Mike Lewis, Luke Zettlemoyer, and Veselin Stoyanov. 2019.
\newblock \href {http://arxiv.org/abs/1907.11692} {Roberta: {A} robustly optimized {BERT} pretraining approach}.
\newblock \emph{CoRR}, abs/1907.11692.

\bibitem[{Mao and Nakagawa(2023)}]{mao-nakagawa-2023-lealla}
Zhuoyuan Mao and Tetsuji Nakagawa. 2023.
\newblock \href {https://doi.org/10.18653/v1/2023.eacl-main.138} {{{LEALLA}}: {{Learning Lightweight Language-agnostic Sentence Embeddings}} with {{Knowledge Distillation}}}.
\newblock In \emph{Proceedings of the 17th {{Conference}} of the {{European Chapter}} of the {{Association}} for {{Computational Linguistics}}}, pages 1886--1894, {Dubrovnik, Croatia}. {Association for Computational Linguistics}.

\bibitem[{Matos~Veliz et~al.(2019)Matos~Veliz, De~Clercq, and Hoste}]{matosveliz-etal-2019-benefits}
Claudia Matos~Veliz, Orphee De~Clercq, and Veronique Hoste. 2019.
\newblock \href {https://doi.org/10.18653/v1/D19-5536} {Benefits of {{Data Augmentation}} for {{NMT-based Text Normalization}} of {{User-Generated Content}}}.
\newblock In \emph{Proceedings of the 5th {{Workshop}} on {{Noisy User-generated Text}} ({{W-NUT}} 2019)}, pages 275--285, {Hong Kong, China}. {Association for Computational Linguistics}.

\bibitem[{Ott et~al.(2019)Ott, Edunov, Baevski, Fan, Gross, Ng, Grangier, and Auli}]{ott-2019-fairseq}
Myle Ott, Sergey Edunov, Alexei Baevski, Angela Fan, Sam Gross, Nathan Ng, David Grangier, and Michael Auli. 2019.
\newblock \href {https://doi.org/10.18653/v1/N19-4009} {fairseq: A fast, extensible toolkit for sequence modeling}.
\newblock In \emph{Proceedings of the 2019 Conference of the North {A}merican Chapter of the Association for Computational Linguistics (Demonstrations)}, pages 48--53, Minneapolis, Minnesota. Association for Computational Linguistics.

\bibitem[{Plank et~al.(2020)Plank, Jensen, and van~der Goot}]{plank-etal-2020-dan}
Barbara Plank, Kristian~N{\o}rgaard Jensen, and Rob van~der Goot. 2020.
\newblock \href {https://doi.org/10.18653/v1/2020.coling-main.583} {{D}a{N}+: {D}anish nested named entities and lexical normalization}.
\newblock In \emph{Proceedings of the 28th International Conference on Computational Linguistics}, pages 6649--6662, Barcelona, Spain (Online). International Committee on Computational Linguistics.

\bibitem[{Reimers and Gurevych(2020)}]{reimers-gurevych-2020-making}
Nils Reimers and Iryna Gurevych. 2020.
\newblock \href {https://doi.org/10.18653/v1/2020.emnlp-main.365} {Making {{Monolingual Sentence Embeddings Multilingual}} using {{Knowledge Distillation}}}.
\newblock In \emph{Proceedings of the 2020 {{Conference}} on {{Empirical Methods}} in {{Natural Language Processing}} ({{EMNLP}})}, pages 4512--4525, {Online}. {Association for Computational Linguistics}.

\bibitem[{Riabi et~al.(2021)Riabi, Sagot, and Seddah}]{riabi-etal-2021-can}
Arij Riabi, Beno{\^i}t Sagot, and Djam{\'e} Seddah. 2021.
\newblock \href {https://doi.org/10.18653/v1/2021.wnut-1.47} {Can {{Character-based Language Models Improve Downstream Task Performances In Low-Resource And Noisy Language Scenarios}}?}
\newblock In \emph{Proceedings of the {{Seventh Workshop}} on {{Noisy User-generated Text}} ({{W-NUT}} 2021)}, pages 423--436, {Online}. {Association for Computational Linguistics}.

\bibitem[{Rosales~N{\'u}{\~n}ez et~al.(2021{\natexlab{a}})Rosales~N{\'u}{\~n}ez, Seddah, and Wisniewski}]{rosalesnunez-etal-2021-understanding}
Jos{\'e}~Carlos Rosales~N{\'u}{\~n}ez, Djam{\'e} Seddah, and Guillaume Wisniewski. 2021{\natexlab{a}}.
\newblock \href {https://aclanthology.org/2021.wnut-1.22} {Understanding the {{Impact}} of {{UGC Specificities}} on {{Translation Quality}}}.
\newblock In \emph{Proceedings of the {{Seventh Workshop}} on {{Noisy User-generated Text}} ({{W-NUT}} 2021)}, pages 189--198, {Online}. {Association for Computational Linguistics}.

\bibitem[{Rosales~N{\'u}{\~n}ez et~al.(2021{\natexlab{b}})Rosales~N{\'u}{\~n}ez, Wisniewski, and Seddah}]{rosales-nunez-etal-2021-noisy}
Jos{\'e}~Carlos Rosales~N{\'u}{\~n}ez, Guillaume Wisniewski, and Djam{\'e} Seddah. 2021{\natexlab{b}}.
\newblock \href {https://doi.org/10.18653/v1/2021.wnut-1.23} {Noisy {UGC} translation at the character level: Revisiting open-vocabulary capabilities and robustness of char-based models}.
\newblock In \emph{Proceedings of the Seventh Workshop on Noisy User-generated Text (W-NUT 2021)}, pages 199--211, Online. Association for Computational Linguistics.

\bibitem[{Samuel and Straka(2021)}]{samuel-straka-2021-ufal}
David Samuel and Milan Straka. 2021.
\newblock \href {https://doi.org/10.18653/v1/2021.wnut-1.54} {{{{\'U}FAL}} at {{MultiLexNorm}} 2021: {{Improving Multilingual Lexical Normalization}} by {{Fine-tuning ByT5}}}.
\newblock In \emph{Proceedings of the {{Seventh Workshop}} on {{Noisy User-generated Text}} ({{W-NUT}} 2021)}, pages 483--492, {Online}. {Association for Computational Linguistics}.

\bibitem[{Sanguinetti et~al.(2020)Sanguinetti, Bosco, Cassidy, {\c{C}}etino{\u{g}}lu, Cignarella, Lynn, Rehbein, Ruppenhofer, Seddah, and Zeldes}]{sanguinetti-etal-2020-treebanking}
Manuela Sanguinetti, Cristina Bosco, Lauren Cassidy, {\"O}zlem {\c{C}}etino{\u{g}}lu, Alessandra~Teresa Cignarella, Teresa Lynn, Ines Rehbein, Josef Ruppenhofer, Djam{\'e} Seddah, and Amir Zeldes. 2020.
\newblock \href {https://aclanthology.org/2020.lrec-1.645} {Treebanking user-generated content: A proposal for a unified representation in {U}niversal {D}ependencies}.
\newblock In \emph{Proceedings of the Twelfth Language Resources and Evaluation Conference}, pages 5240--5250, Marseille, France. European Language Resources Association.

\bibitem[{Schuster and Paliwal(1997)}]{schuster-paliwal-1997-bidirectional}
M.~Schuster and K.K. Paliwal. 1997.
\newblock \href {https://doi.org/10.1109/78.650093} {Bidirectional recurrent neural networks}.
\newblock \emph{IEEE Transactions on Signal Processing}, 45(11):2673--2681.

\bibitem[{Seddah et~al.(2012)Seddah, Sagot, Candito, Mouilleron, and Combet}]{seddah-etal-2012-french}
Djam{\'e} Seddah, Benoit Sagot, Marie Candito, Virginie Mouilleron, and Vanessa Combet. 2012.
\newblock \href {https://aclanthology.org/C12-1149} {The {F}rench {S}ocial {M}edia {B}ank: a treebank of noisy user generated content}.
\newblock In \emph{Proceedings of {COLING} 2012}, pages 2441--2458, Mumbai, India. The COLING 2012 Organizing Committee.

\bibitem[{Tan et~al.(2023)Tan, Heffernan, Schwenk, and Koehn}]{tan-etal-2023-multilingual}
Weiting Tan, Kevin Heffernan, Holger Schwenk, and Philipp Koehn. 2023.
\newblock \href {https://doi.org/10.18653/v1/2023.eacl-main.108} {Multilingual representation distillation with contrastive learning}.
\newblock In \emph{Proceedings of the 17th Conference of the European Chapter of the Association for Computational Linguistics}, pages 1477--1490, Dubrovnik, Croatia. Association for Computational Linguistics.

\bibitem[{Tang et~al.(2022)Tang, Kocyigit, and Wijaya}]{tang-etal-2022-augcse}
Zilu Tang, Muhammed~Yusuf Kocyigit, and Derry~Tanti Wijaya. 2022.
\newblock \href {https://aclanthology.org/2022.aacl-main.30} {{{AugCSE}}: {{Contrastive Sentence Embedding}} with {{Diverse Augmentations}}}.
\newblock In \emph{Proceedings of the 2nd {{Conference}} of the {{Asia-Pacific Chapter}} of the {{Association}} for {{Computational Linguistics}} and the 12th {{International Joint Conference}} on {{Natural Language Processing}} ({{Volume}} 1: {{Long Papers}})}, pages 375--398, {Online only}. {Association for Computational Linguistics}.

\bibitem[{Tay et~al.(2022)Tay, Tran, Ruder, Gupta, Chung, Bahri, Qin, Baumgartner, Yu, and Metzler}]{tay-etal-2021-charformer}
Yi~Tay, Vinh~Q. Tran, Sebastian Ruder, Jai~Prakash Gupta, Hyung~Won Chung, Dara Bahri, Zhen Qin, Simon Baumgartner, Cong Yu, and Donald Metzler. 2022.
\newblock \href {https://openreview.net/forum?id=JtBRnrlOEFN} {Charformer: Fast character transformers via gradient-based subword tokenization}.
\newblock In \emph{Proceedings of the The Tenth International Conference on Learning Representations}, Virtual.

\bibitem[{{van der Goot}(2019)}]{vandergoot-2019-indepth}
Rob {van der Goot}. 2019.
\newblock \href {https://doi.org/10.18653/v1/D19-5515} {An {{In-depth Analysis}} of the {{Effect}} of {{Lexical Normalization}} on the {{Dependency Parsing}} of {{Social Media}}}.
\newblock In \emph{Proceedings of the 5th {{Workshop}} on {{Noisy User-generated Text}} ({{W-NUT}} 2019)}, pages 115--120, {Hong Kong, China}. {Association for Computational Linguistics}.

\bibitem[{{van der Goot} et~al.(2018){van der Goot}, {van Noord}, and {van Noord}}]{vandergoot-etal-2018-taxonomy}
Rob {van der Goot}, Rik {van Noord}, and Gertjan {van Noord}. 2018.
\newblock \href {https://aclanthology.org/L18-1109} {A {{Taxonomy}} for {{In-depth Evaluation}} of {{Normalization}} for {{User Generated Content}}}.
\newblock In \emph{International {{Conference}} on {{Language Resources}} and {{Evaluation}}}, Miyazaki, Japan. European Language Resources Association (ELRA).

\bibitem[{Vaswani et~al.(2017)Vaswani, Shazeer, Parmar, Uszkoreit, Jones, Gomez, Kaiser, and Polosukhin}]{vaswani-etal-2017-attention}
Ashish Vaswani, Noam Shazeer, Niki Parmar, Jakob Uszkoreit, Llion Jones, Aidan~N. Gomez, Lukasz Kaiser, and Illia Polosukhin. 2017.
\newblock \href {http://arxiv.org/abs/1706.03762} {Attention is all you need}.

\bibitem[{Wang et~al.(2022)Wang, Yang, Huang, Jiao, Yang, Jiang, Majumder, and Wei}]{wang-etal-2022-text}
Liang Wang, Nan Yang, Xiaolong Huang, Binxing Jiao, Linjun Yang, Daxin Jiang, Rangan Majumder, and Furu Wei. 2022.
\newblock \href {https://doi.org/10.48550/ARXIV.2212.03533} {Text embeddings by weakly-supervised contrastive pre-training}.
\newblock \emph{CoRR}, abs/2212.03533.

\bibitem[{Yan et~al.(2021)Yan, Li, Wang, Zhang, Wu, and Xu}]{yan-etal-2021-consert}
Yuanmeng Yan, Rumei Li, Sirui Wang, Fuzheng Zhang, Wei Wu, and Weiran Xu. 2021.
\newblock \href {https://doi.org/10.18653/v1/2021.acl-long.393} {{{ConSERT}}: {{A Contrastive Framework}} for {{Self-Supervised Sentence Representation Transfer}}}.
\newblock In \emph{Proceedings of the 59th {{Annual Meeting}} of the {{Association}} for {{Computational Linguistics}} and the 11th {{International Joint Conference}} on {{Natural Language Processing}} ({{Volume}} 1: {{Long Papers}})}, pages 5065--5075, {Online}. {Association for Computational Linguistics}.

\end{thebibliography}

\begin{thebibliography}{7}
\expandafter\ifx\csname natexlab\endcsname\relax\def\natexlab#1{#1}\fi

\bibitem[{Bawden and Sagot(2023)}]{ROCSMT}
Bawden, Rachel and Sagot, Beno{\^\i}t. 2023.
\newblock \href {https://doi.org/10.18653/v1/2023.wmt-1.21} {\emph{{R}o{CS}-{MT}: Robustness Challenge Set for Machine Translation}}.
\newblock Association for Computational Linguistics.

\bibitem[{Michel and Neubig(2018)}]{MTNT}
Michel, Paul and Neubig, Graham. 2018.
\newblock \href {https://doi.org/10.18653/v1/D18-1050} {\emph{{MTNT}: A Testbed for Machine Translation of Noisy Text}}.
\newblock Association for Computational Linguistics.

\bibitem[{Muennighoff et~al.(2023)Muennighoff, Tazi, Magne, and Reimers}]{MTEB}
Muennighoff, Niklas and Tazi, Nouamane and Magne, Loic and Reimers, Nils. 2023.
\newblock \href {https://doi.org/10.18653/v1/2023.eacl-main.148} {\emph{{MTEB}: Massive Text Embedding Benchmark}}.
\newblock Association for Computational Linguistics.

\bibitem[{{NLLB Team} et~al.(2022){NLLB Team}, Costa-jussà, Cross, Çelebi, Elbayad, Heafield, Heffernan, Kalbassi, Lam, Licht, Maillard, Sun, Wang, Wenzek, Youngblood, Akula, Barrault, Mejia~Gonzalez, Hansanti, Hoffman, Jarrett, Ram~Sadagopan, Rowe, Spruit, Tran, Andrews, Ayan, Bhosale, Edunov, Fan, Gao, Goswami, Guzm\'an, Koehn, Mourachko, Ropers, Saleem, Schwenk, and Wang}]{NLLB}
{NLLB Team} and Costa-jussà, Marta R. and Cross, James and Çelebi, Onur and Elbayad, Maha and Heafield, Kenneth and Heffernan, Kevin and Kalbassi, Elahe and Lam, Janice and Licht, Daniel and Maillard, Jean and Sun, Anna and Wang, Skyler and Wenzek, Guillaume and Youngblood, Al and Akula, Bapi and Barrault, Lo\"ic and Mejia Gonzalez, Gabriel and Hansanti, Prangthip and Hoffman, John and Jarrett, Semarley and Ram Sadagopan, Kaushik and Rowe, Dirk and Spruit, Shannon and Tran, Chau and Andrews, Pierre and Ayan, Necip Fazil and Bhosale, Shruti and Edunov, Sergey and Fan, Angela and Gao, Cynthia and Goswami, Vedanuj and Guzm\'an, Francisco and Koehn, Philipp and Mourachko, Alexandre and Ropers, Christophe and Saleem, Safiyyah and Schwenk, Holger and Wang, Jeff. 2022.
\newblock \emph{No Language Left Behind: Scaling Human-Centered Machine Translation}.
\newblock FAIR (META).

\bibitem[{Ortiz~Su{\'a}rez et~al.(2019)Ortiz~Su{\'a}rez, Sagot, and Romary}]{OSCAR}
Ortiz Su{\'a}rez, Pedro Javier and Sagot, Beno\^it and Romary, Laurent. 2019.
\newblock \href {https://doi.org/10.14618/ids-pub-9021} {\emph{Asynchronous pipelines for processing huge corpora on medium to low resource infrastructures}}.
\newblock Leibniz-Institut f{\"u}r Deutsche Sprache.

\bibitem[{Rosales~N{\'u}{\~n}ez et~al.(2019)Rosales~N{\'u}{\~n}ez, Seddah, and Wisniewski}]{PFSMB}
Rosales N{\'u}{\~n}ez, Jos{\'e} Carlos and Seddah, Djam{\'e} and Wisniewski, Guillaume. 2019.
\newblock \href {https://aclanthology.org/W19-6101} {\emph{Comparison between {NMT} and {PBSMT} Performance for Translating Noisy User-Generated Content}}.
\newblock Link{\"o}ping University Electronic Press.

\bibitem[{{van der Goot} et~al.(2021){van der Goot}, Ramponi, Zubiaga, Plank, Muller, San Vicente~Roncal, Ljube{\v s}i{\'c}, {\c C}etino{\u g}lu, Mahendra, {\c C}olako{\u g}lu, Baldwin, Caselli, and Sidorenko}]{MultiLexNorm2021}
{van der Goot}, Rob and Ramponi, Alan and Zubiaga, Arkaitz and Plank, Barbara and Muller, Benjamin and San Vicente Roncal, I{\~n}aki and Ljube{\v s}i{\'c}, Nikola and {\c C}etino{\u g}lu, {\"O}zlem and Mahendra, Rahmad and {\c C}olako{\u g}lu, Talha and Baldwin, Timothy and Caselli, Tommaso and Sidorenko, Wladimir. 2021.
\newblock \href {https://aclanthology.org/2021.wnut-1.55.pdf} {\emph{{{MultiLexNorm}}: {{A Shared Task}} on {{Multilingual Lexical Normalization}}}}.
\newblock {Association for Computational Linguistics}.

\end{thebibliography}

\section{Language Resource References}
\label{lr:ref}
\bibliographystylelanguageresource{lrec-coling2024-natbib}

\appendix

\section*{Appendices}

\section{Transformations for Artificial UGC Generation} \label{appendix:transformations}

Below is the detailed list of transformations selected from NL-Augmenter for artificial UGC generation.
\begin{enumerate}
 \item \texttt{abr1} \textit{(abbreviation\_transformation)}:\footnote{Name of the transformation module in NL-Augmenter.} replaces words or phrases with their abbreviated counterpart using a web-scraped slang dictionary (with default probability $p=0.1$)
 \item \texttt{abr2} \textit{(insert\_abbreviation)}: replaces words or phrases with their abbreviated counterpart from a list of common generic and social media abbreviations
 \item \texttt{abr3} \textit{(replace\_abbreviation\_and\_acronyms)}: swaps the abbreviated and expanded forms of words and phrases from a list of common abbreviations and acronyms in business communications
 \item \texttt{cont} \textit{(contraction\_expansions)}: swaps commonly used contractions and expansions, \textit{e.g.~I am $\leftrightarrow$ I'm}
 \item \texttt{dysl} \textit{(dyslexia\_words\_swap)}: replaces words with their counterparts from a list of frequently misspelled words for dyslexia, \textit{e.g.~lose $\leftrightarrow$ loose}
 \item \texttt{fing} \textit{(butter\_fingers\_perturbation)}: swaps letters with one of their QWERTY keyboard neighbours ($p=0.05$)
 \item \texttt{homo} \textit{(close\_homophones\_swap)}: replaces words with one of their homophones ($p=0.5$), \textit{e.g.~there $\leftrightarrow$ their}
 \item \texttt{leet} \textit{(leet\_letters)}: replaces letters with their Leet\footnote{\url{https://en.wikipedia.org/wiki/Leet}} equivalents ($p=0.1$), \textit{e.g.~love $\rightarrow$ l0V3}
 \item \texttt{slng} \textit{(slangificator)}: replaces words (in particular, nouns, adjectives, and adverbs) with their corresponding slang from a dictionary of English slang and colloquialisms
 \item \texttt{spac} \textit{(whitespace\_perturbation)}: adds or remove a whitespace at random positions ($p_{add}=0.05, p_{remove}=0.1$)
 \item \texttt{spel} \textit{(replace\_spelling)}: replaces words with their counterparts from corpora of frequently misspelled words ($p=0.2$)
 \item \texttt{week} \textit{(weekday\_month\_abbreviation)}: abbreviates or expands the names of months and weekdays, \textit{e.g.~Mon. $\leftrightarrow$ Monday}
\end{enumerate}
\begin{table*}[!ht]
    \centering
    \small
    \setlength\tabcolsep{0pt}
    \begin{tabular*}{\textwidth}{@{\extracolsep{\fill}}lcccccccccccccc@{}}
    \toprule
    UGC type & abr1 & abr2 & abr3 & cont & dysl & fing & homo & leet & slng & spac & spel & week & mix\_all \\ \midrule
    LASER & 0.15 & 0.30 & \textbf{0.00} & 0.10 & 0.10 & 0.35 & 0.11 & 10.35 & 0.10 & 0.36 & 0.16 & \textbf{0.00} & 0.38 \\
    RoLASER & \textbf{0.00} & \textbf{0.00} & \textbf{0.00} & \textbf{0.00} & \textbf{0.00} & \textbf{0.00} & \textbf{0.00} & \textbf{0.00}** & \textbf{0.00} & \textbf{0.00} & \textbf{0.00} & \textbf{0.00} & \textbf{0.00}** \\
   c-RoLASER & \textbf{0.00} & \textbf{0.00} & \textbf{0.00} & \textbf{0.00} & \textbf{0.00} & \textbf{0.00} & \textbf{0.00} & \textbf{0.00}** & \textbf{0.00} & \textbf{0.00} & \textbf{0.00} & \textbf{0.00} & \textbf{0.00}** \\
   \bottomrule
    \end{tabular*}
    \caption{xSIM scores for all models on UGC$\rightarrow$standard English bitext from each UGC type of FLORES\textsuperscript{$\dagger$} devtest data, averaged across 10 data generation seeds. The best score for each type is in \textbf{bold}. **: $p<0.001$, statistical significance with respect to LASER's scores.}
    \label{tab:artificial-ugc-student-xsim}
\end{table*}

\begin{figure}[!ht]
 \centering
 \includegraphics[width=0.5\textwidth]{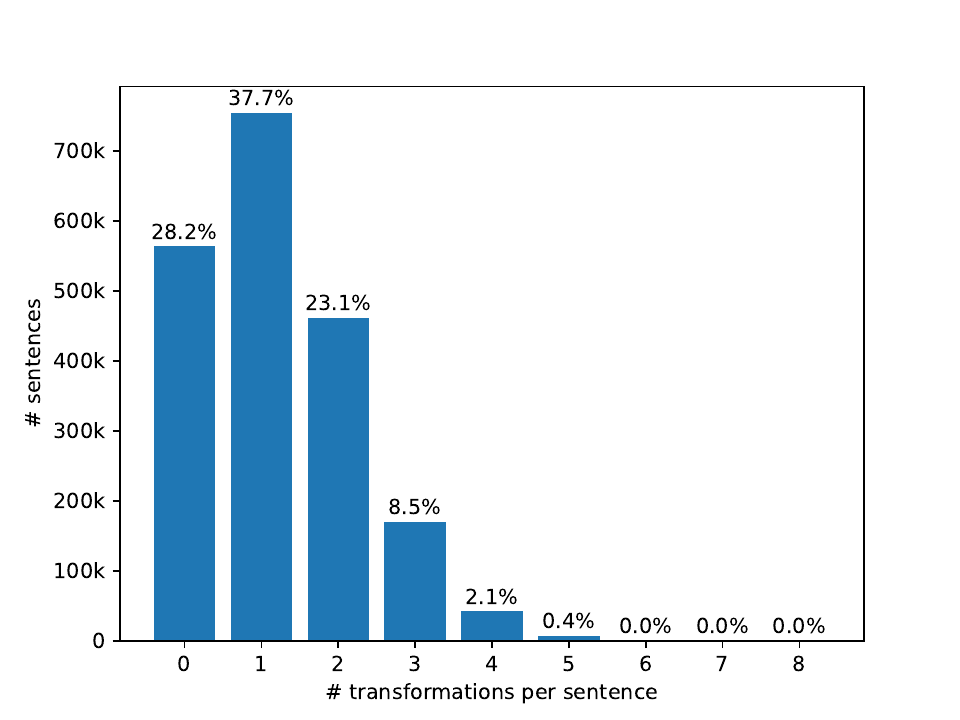}
 \caption{Distribution of transformations obtained by applying \texttt{mix\_all} on 2M training sentences.}
 \label{fig:distribution}
\end{figure}

We also implement a \texttt{mix\_all} transformation that combines perturbations from some of the 12 transformations. Firstly, a subset of the transformations is uniformly selected with probability $p_{all}=0.1$. Then they are shuffled, ensuring that they are not always applied in the same order. Lastly, for the transformations that depend on a probability parameter $p$, let $p_{d}$ denote its default value. The value of $p$ is randomly selected between \{$\frac{1}{2}p_{d}$, $p_{d}$, $\frac{3}{2}p_{d}$\}, with probabilities of \{$\frac{1}{4}$, $\frac{1}{2}$, $\frac{1}{4}$\} respectively. A different random seed is used for each transformation. Figure~\ref{fig:distribution} illustrates the distribution of the number of perturbations applied to each sentence as a result of executing the \texttt{mix\_all} transformation on 2 million training sentences from the OSCAR dataset. 

\begin{figure}[!th]
 \centering
 \includegraphics[width=0.485\textwidth]{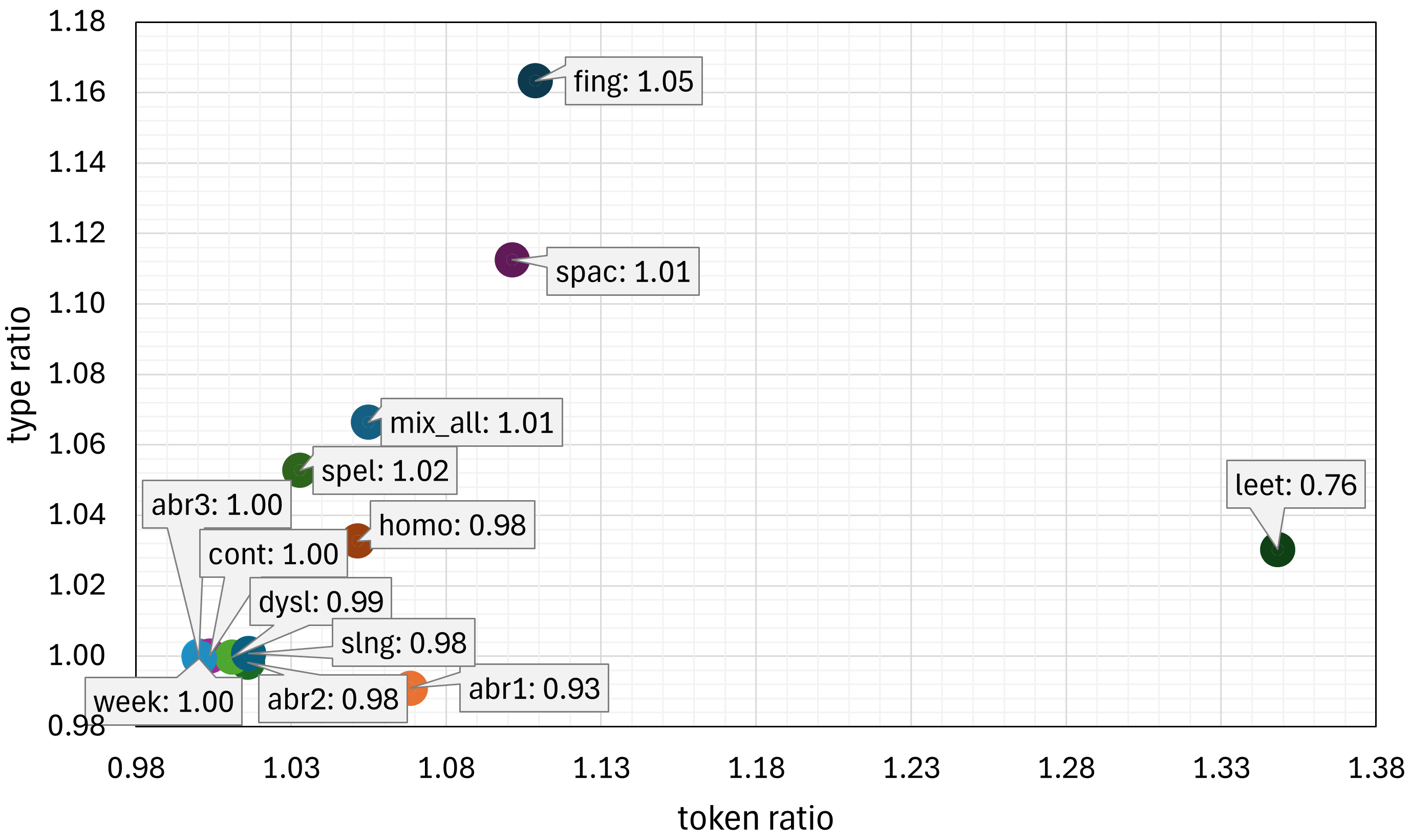}
 \caption{Visualisation of UGC phenomena of the FLORES\textsuperscript{$\dagger$} devtest by their type and token ratios. The data point labels indicate TTR ratios. All ratios are with respect to the standard English text.}
 \label{fig:ttr}
\end{figure}

All these transformations produce artificial UGC texts with varying levels of non-standardness. Figure~\ref{fig:ttr} illustrates the ratios of number of types, number of tokens and TTR of the FLORES\textsuperscript{$\dagger$} devtest texts generated by each transformation with respect to the standard English text. The perturbations with the highest and lowest TTR ratios are \texttt{fing} and \texttt{leet}, respectively. \texttt{fing} also has the highest type ratio while \texttt{leet} has the highest token ratio. Both transformations perform character-level substitutions that shatter LASER's SentencePiece tokenisation. \texttt{spac} also has a high type ratio as a result of inserting and deleting whitespaces. In theory, the closer a transformation is to the lower-left corner of the plot, the more standard-like the UGC text is. For instance, \texttt{abr3}, \texttt{cont} and \texttt{week} fall into this category with all three ratios equal to $1.00$. Conversely, the farther the transformation is from the lower-left corner, the more non-standard it is (and therefore more challenging for LASER).

\section{Comparison of Cosine Distance, xSIM and xSIM++ across Languages}
\label{appendix:quantiles}

\begin{figure}[!ht]
 \centering
 \includegraphics[width=0.5\textwidth]{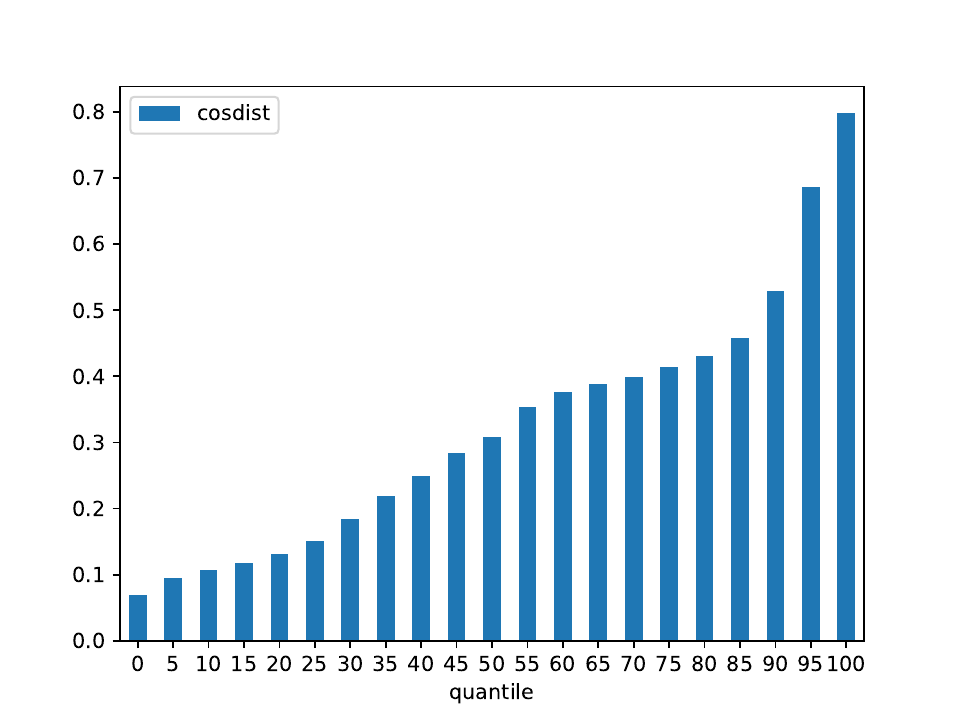}
 \caption{Quantiles of average pairwise cosine distance on FLORES devtest for all 199 xx$\rightarrow$English language pairs.}
 \label{fig:cos-quantiles}
\end{figure}

\begin{figure}[!ht]
 \centering
 \includegraphics[width=0.5\textwidth]{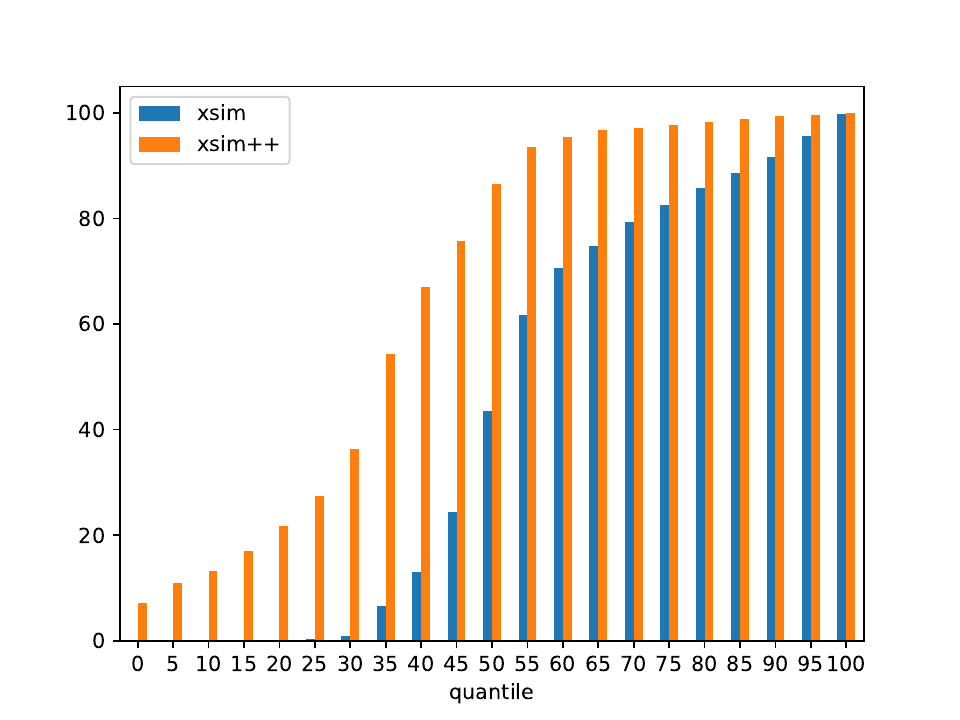}
 \caption{Quantiles of xSIM and xSIM++ scores on FLORES devtest for all 199 xx$\rightarrow$English language pairs.}
 \label{fig:quantiles}
\end{figure}

The FLORES dataset has $n$-way parallel texts in 200 languages. We produce LASER embeddings of the devtest and compute average pairwise cosine distance, xSIM and xSIM++ for all 199 xx-English language pairs. Figure~\ref{fig:cos-quantiles} shows the quantiles of the cosine distance, while Figure~\ref{fig:quantiles} shows those of xSIM and xSIM++. The minimum values (or 0\textsuperscript{th} percentiles) are 0.07, 0\% and 7.21\% for cosine distance, xSIM and xSIM++ respectively.

Notably, Figure~\ref{fig:quantiles} supports the observation made by \citet{chen-etal-2023-xsim} that the xSIM scores for many language pairs ``quickly saturate at 0\%''. Indeed, the xSIM value remains at 0 until the 20\% quantile (20\textsuperscript{th} percentile). This means that for the top 20\% language pairs, LASER has a perfect xSIM score in aligning the sentences. We see that xSIM++ is a better metric because it is not easy to get a perfect score. It is therefore deemed more ``challenging''. 


\section{xSIM Scores on Artificial UGC}
\label{appendix:xsim}

Table~\ref{tab:artificial-ugc-student-xsim} shows the xSIM scores of the three models on the artificial UGC texts from FLORES\textsuperscript{$\dagger$} \mbox{devtest}. Both RoLASER and \crolaser\ get a consistent score of zero across all UGC types. As it has already been stated that xSIM is not challenging enough on FLORES (see Appendix~\ref{appendix:quantiles}), these results are not informative enough to make further conclusions on their performance, other than that they improve on LASER's.

\end{document}